\documentclass[preprint,12pt]{elsarticle}
\usepackage[letterpaper,margin=2.5cm]{geometry}

\usepackage{amssymb}
\usepackage[T1]{fontenc}
\usepackage{lmodern}
\usepackage{amsmath}
\usepackage{booktabs}
\usepackage{comment}
\usepackage{afterpage}
\usepackage{balance}
\usepackage{graphicx}
\usepackage{IEEEtrantools}  
\usepackage{caption}  
\usepackage{subcaption} 
\usepackage{pdflscape}
\usepackage{longtable}
\usepackage{booktabs}
\usepackage{lscape}
\usepackage{adjustbox}
\usepackage{rotating}
\usepackage{float}
\usepackage{afterpage}
\usepackage{hyperref}
\usepackage{multirow}
\usepackage{arydshln} 

\usepackage{xcolor}
\usepackage[normalem]{ulem}
\usepackage{tabularx}
\usepackage{array}

\newcommand{\wilcoxonpvalue}[2]{\makebox[5.5em][c]{#1\makebox[0pt][l]{#2}}}

\journal{Neurocomputing}

\begin{document}
\begin{@twocolumnfalse}  
\begin{frontmatter}



\title{Assessing the impact of dimensionality reduction on clustering performance — a systematic study}


\author{
Ousmane Assani Amate\textsuperscript{a},
Elyes Lounissi\textsuperscript{a},
Mohammadreza Bakhtyari\textsuperscript{a},
Émilie Roy\textsuperscript{a},
Roman Sarrazin-Gendron\textsuperscript{a},
Vladimir Makarenkov\textsuperscript{a,b} \\[1em]
\textsuperscript{a}Université du Québec à Montréal, Montreal, Quebec, Canada\\
\textsuperscript{b}Mila - Quebec AI Institute, Montreal, Quebec, Canada
}






\begin{abstract}
Dimensionality reduction is a critical preprocessing step for clustering high-dimensional data, yet comprehensive evaluation of its impact across diverse methods and data types remains limited. In this study, we systematically assess the influence of five dimensionality reduction techniques — Principal Component Analysis (PCA), Kernel Principal Component Analysis (Kernel PCA), Variational Autoencoder (VAE), Isometric Mapping (Isomap), and Multidimensional Scaling (MDS) — on the performance of four popular clustering algorithms — $k$-means, Agglomerative Hierarchical Clustering (AHC), Gaussian Mixture Models (GMM), and Ordering Points to Identify the Clustering Structure (OPTICS). We evaluate clustering quality using the Adjusted Rand Index (ARI), comparing results without and with dimensionality reduction at different reduction levels recommended in the literature (i.e., $k-1$, where $k$ is the number of clusters, and 25\% and 50\% of the original number of dimensions).
Our findings underscore the importance of a careful selection of the dimensionality reduction technique and the dimensionality reduction level that should be tailored to intrinsic data geometry and clustering algorithm(s) under consideration.

\end{abstract}



\begin{keyword}
Clustering \sep Dimensionality Reduction \sep Unsupervised Learning


\end{keyword}

\end{frontmatter}
\end{@twocolumnfalse}

\section{Introduction}\label{sec:intro}

Clustering is a fundamental technique in unsupervised machine learning that groups similar data points into clusters based on inherent patterns or similarities. Unlike supervised learning, it operates without labels. Clustering is central to exploratory analysis in various domains such as finance, biomedicine, and computer vision \cite{ezugwu2022comprehensiv,oyewole2023data,tahiri2018new,zhou2024survey}. The main clustering approaches include centroid-based, probabilistic, hierarchical, graph-based, and density-based clustering \cite{wani2024comprehensive}.

High-dimensional data pervades modern machine learning tasks, yet clustering performance often collapses under the "curse of dimensionality", as data points become nearly equidistant and classical notions of similarity lose discriminative power \cite{thrun2021distance}. Dimensionality reduction offers a promising avenue by projecting data into a lower-dimensional representation that mitigates sparsity and noise while preserving essential data structure. Such projections can reveal latent manifolds, filter redundancy, remove noisy features, and enhance the separability of clusters, making dimensionality reduction a common preprocessing step in unsupervised analyses \cite{jia2022feature}.

Dimensionality reduction methods can be broadly grouped into three methodological families, each offering distinct advantages for clustering \cite{niu2025exploring}. Projection-based methods such as PCA \cite{pearson1901liii} and its nonlinear kernel variant, Kernel PCA \cite{scholkopf1998kpca}, seek low-dimensional subspaces that respectively capture maximal variance based on linear and non-linear relationships between variables. Geometry-based methods, including MDS \cite{kruskal1964multidimensional}, Isomap \cite{tenenbaum2000isomap}, Locally Linear Embedding  \cite{sam2000lle}, Laplacian Eigenmaps \cite{belkin2003le}, Diffusion Maps \cite{coifman2006}, t-distributed Stochastic Neighbor Embedding (t-SNE) \cite{vandermaaten2008tsne}, and Uniform Manifold Approximation and Projection (UMAP) \cite{healy2024uniform}, preserve local or global geometric relationships in the embedding, often revealing nonlinear structures invisible to linear projections. Deep learning–based approaches, such as Autoencoders \cite{hinton2006ae} and Variational Autoencoders (VAEs) \cite{kingma2013vae}, learn highly nonlinear transformations through neural networks, enabling scalable dimensionality reduction for large, complex, and multimodal datasets.

Although dimensionality reduction and clustering have both been extensively studied, most previous work treats them separately \cite{ray2021various,rodriguez2019clustering} or explores their specific combinations \cite{ding2004kmeanspca}, often focusing on a single dimensionality reduction method or clustering algorithm \cite{allaoui2020umap,alkhayrat2020comparative}, considering domain-specific applications \cite{rovira2022reactive,sun2024genomics}, or using dimensionality reduction primarily for visualization \cite{xia2022vis}. Moreover, many experimental studies are performed in supervised or semi-supervised settings \cite{ayesha2020overview}, leaving a gap in systematic, large-scale, unsupervised comparison. To the best of our knowledge, no domain-agnostic study has jointly examined diverse dimensionality reduction families (linear, manifold, and deep learning) across multiple clustering algorithms, dataset types, and dimensionality reduction levels, within a controlled and reproducible framework.

Our work addresses this gap by providing a statistically grounded, large-scale comparison of dimensionality reduction techniques used prior to data clustering. Precisely, we assess the performance of five dimensionality reduction methods (PCA, Kernel PCA, VAE, Isomap, and MDS) at three dimensionality reduction levels: $k-1$ (where $k$ is the number of clusters), and 25\% and 50\% of the original number of dimensions. These methods are applied in combination with four well-known clustering algorithms: $k$-means, Agglomerative Hierarchical Clustering (AHC), Gaussian Mixture Models (GMM), and Ordering Points to Identify the Clustering Structure (OPTICS). Using both synthetic and real-world data, we quantify clustering quality with an external metric (ARI), enabling direct and fair clustering performance comparison. Our key contributions are threefold:

\begin{itemize}
    \item \textbf{Comprehensive evaluation:} Systematic comprehensive comparison of five dimensionality reduction methods used in combination with four clustering algorithms across different dimensionality reduction levels, conducted under a statistically controlled experimental design.
    \item \textbf{Quantitative benchmarking:} Statistically validated performance analysis using Adjusted Rand Index (ARI) and the Wilcoxon signed-rank test over diverse synthetic datasets and real-world benchmarks.
    \item \textbf{Practical guidance:} Actionable recommendations for selecting suitable dimensionality reduction method/clustering algorithm pairings based on data characteristics and clustering objectives, supported by open-source code and reproducible experiments.
\end{itemize}

The rest of the paper is organized as follows: Section \ref{sec:related_work} reviews related work in the field. Section \ref{sec:methodology} describes synthetic and real-world data used in our computational experiments, then presents dimensionality reduction methods, clustering algorithms, evaluation metrics, and experimental setup. Section \ref{sec:results} outlines our experimental results obtained for synthetic and real-world datasets considered. Section \ref{sec:discussion} discusses the outcomes of our experiments. Finally, Section \ref{sec:conclusion} concludes the study and outlines directions for future research.

\section{Related Work} \label{sec:related_work}

In high-dimensional spaces, distance distributions concentrate and standard metrics (e.g., Euclidean distance) lose their discriminative power, degrading similarity search and clustering quality. This motivates the use of dimensionality reduction prior to clustering \cite{herrmann2024enhancing,aggarwal2001surprising,beyer1999meaningful}. A large body of work has investigated pipelines that combine dimensionality reduction methods with clustering algorithms. These studies can be typically divided into three categories, where: (i) Both dimensionality reduction method and clustering algorithm are fixed, (ii) Clustering algorithm is fixed and dimensionality reduction methods vary, and (iii) Dimensionality reduction method is fixed and clustering algorithms vary. 

A canonical example of the first category of work is the combination of PCA and $k$-means, where theoretical results connect principal components to relaxed cluster indicator vectors \cite{ding2004kmeanspca}. Mixtures of probabilistic principal component analyzers can integrate PCA-like latent spaces directly into Gaussian mixtures, effectively performing joint dimensionality reduction and clustering \cite{tipping1999mppca}. In geometry-based and manifold-based approaches, Isomap or MDS dimensionality reduction technique is often followed by $k$-means or GMM, providing practical gains in clustering quality
\cite{tseng2023isomapkmeans}.

Several studies have reported that the choice of dimensionality reduction technique substantially influences clustering performance of a given (fixed) clustering algorithm. In the case of $k$-means clustering, Alkhayrat et al. \cite{alkhayrat2020comparative} showed that a deep autoencoder latent space constitutes a better clustering ground than a PCA-based latent space or the original feature space, with the number of retained latent features strongly influencing clustering quality. Extending beyond $k$-means, Roh et al. \cite{roh2025kche} investigated the performance of GMM, systematically varying dimensionality reduction methods (i.e., considering PCA, ICA, Isomap, LLE, t-SNE, and UMAP). The authors used ARI and the Normalized Mutual Information (NMI) index to evaluate the effectiveness of  dimensionality reduction for subsequent clustering. They concluded that UMAP demonstrated notable results, especially with sparse and noisy data, while offering a decent robustness performance with out-of-sample test data.

Different studies involved experiments in which dimensionality reduction method was held fixed, but clustering algorithms varied. For instance, using UMAP embedding, Allaoui et al. \cite{allaoui2020umap} reported accuracy gains of up to 60\% when $k$-means, Agglomerative Clustering, GMM, and HDBSCAN operated on the low-dimension feature space provided by UMAP. Remarkably, a series of experiments on image datasets showed that UMAP allowed each of the clustering algorithms considered to improve its performance on each evaluated dataset. Furthermore, Dalmaijer et al. \cite{dalmaijer2022power} carried out a simulation-driven investigation with UMAP and found that $k$-means and Agglomerative Clustering performed similarly with or without dimensionality reduction, whereas HDBSCAN benefited markedly from the use of UMAP.

Beyond general pipelines, domain-specific workflows further illustrate the value of combining dimensionality reduction and clustering. In the field of biomedicine, Lötsch et al. \cite{loetsch2024comparative} evaluated the impact of six dimensionality reduction methods (i.e., PCA, ICA, Isomap, MDS, t-SNE, and UMAP) on the performance of five clustering algorithms (i.e., $k$-means, $k$-medoids, single and average linkage, and Ward's method) using nine synthetic and five real-world datasets. The authors concluded that no dimensionality reduction method/clustering algorithm pairing was able to recover consistently true biomedical image classification. Lötsch et al. also pointed out that PCA often outperformed or matched the results of the neighborhood-based methods (i.e., UMAP and t-SNE) and the manifold learning technique (Isomap). They concluded that the selection of dimensionality reduction method should be data specific to establish a tailored approach to data projection and clustering in biomedical analysis. In the field of reactive flow physics, Rovira et al. \cite{rovira2022reactive} proposed a three-step workflow, first applying a nonlinear dimensionality reduction method, then performing unsupervised clustering, and finally conducting feature correlation analysis, to reveal meaningful patterns and structures in the data. Sun et al. \cite{sun2024genomics} have recently surveyed common practices in single-cell and spatial transcriptomics, such as the construction of PCA-based graphs for Louvain or Leiden clustering or the use of manifold learning methods for visualization, highlighting that effective clustering workflows must be adapted to specific constraints of each data modality. These field-oriented studies provide valuable insights but remain clearly domain-tuned with their specific protocols, including data pre-processing, feature reduction methods and levels, clustering algorithms, and evaluation metrics, varying widely.

Overall, three consistent empirical patterns emerge: (i) Nonlinear and deep embeddings (e.g., UMAP and autoencoders) often improve clustering stability and accuracy relative to linear PCA when the underlying structure is nonlinear \cite{alkhayrat2020comparative,allaoui2020umap,dalmaijer2022power}, (ii) Density-based clustering algorithms (e.g., HDBSCAN) provide the largest relative gains after manifold dimensionality reduction, from which partitioning and hierarchical clustering algorithms benefit more modestly \cite{dalmaijer2022power,allaoui2020umap}, and (iii) Dimensionality reduction level itself is consequential, with clustering performance varying substantially as the target number of dimensions changes \cite{hozumi2021UMAP}. 

Nevertheless, a context-free, systematically controlled dimensionality reduction method/clustering algorithm selection benchmark is still lacking. To respond to this need for further research, we conduct a comprehensive comparison of individual impacts of five dimensionality reduction techniques (i.e., PCA, Kernel PCA, VAE, Isomap, and MDS) on four clustering algorithms ($k$-means, Agglomerative Hierarchical Clustering, GMM, and OPTICS). Each dimensionality reduction method/clustering algorithm pairing is evaluated across three different dimensionality reduction levels recommended in the literature to identify the most effective pairings across various synthetic and real-world benchmarks considered in our study. 
We did not consider here the t-SNE and UMAP dimensionality reduction techniques since both are primarily designed for exploratory visualization; their stochastic embeddings and strong hyperparameter sensitivity may pose a problem for the results reproducibility and fair cross-method comparison \cite{kobak2021initialization}. 

\section{Methodology} \label{sec:methodology}

\subsection{Benchmark data used in the experiments} \label{subsec:datasets}

\subsubsection{Synthetic data} \label{subsubsec:synthetic_datag}

To systematically evaluate the impact of dimensionality reduction on clustering performance, we constructed a diverse suite of synthetic datasets designed to capture a broad range of geometric structures, noise levels, and dimensionalities. These datasets support controlled benchmarking providing different challenging scenarios for clustering algorithms, including high-dimensional complexity, nonlinearity, and varying degrees of cluster overlap. 

First, to assess the methods' performances on nonlinearly separable structures, we generated classical Circle and Moon datasets using the \textit{make\_circles} (Circles) and \textit{make\_moons} (Moons) functions from \textit{scikit-learn} that were supplemented with controlled transformations (stretching, rotations, and translations) to increase structural data diversity. The two-cluster Circles datasets consisted of nested circular structures with $N_c = 1,000$ objects per cluster and the \textit{factor} parameter set to 0.5, ensuring that the inner circle's radius was half that of the outer circle. The five-cluster Circles datasets comprised five concentric rings with $N_c = 400$ objects per cluster. The rings were scaled radially by factors of 1.0, 2.0, 3.5, 5.0, and 7.0, ensuring distinct separations between them. 
The Moons datasets also consisted of two and five-cluster crescent-shaped structures with $n = 2,000$ objects per dataset. The clusters were generated by applying controlled stretching (with factors of 1.0 and 1.5), rotations (with $\pm$160°, $\pm$10°, and 180°), and translations (with x-shifts of $\pm$2 to $\pm$4 and y-shifts of 1.0, 1.2, and 1.5) to create crescent-like arrangements. To enable high-dimensional evaluation, all originally generated Circle and Moon two-dimensional datasets were embedded into 10, 50, and 200 dimensions via Gaussian Random Projection, approximately preserving pairwise distances as per the Johnson–Lindenstrauss lemma \cite{frankl1988johnson}.

Third, we used the data generation method proposed by Rodriguez et al. \cite{rodriguez2019clustering}, allowing one to obtain normally distributed synthetic data with systematic control over cluster separability and feature covariance. We hereafter denote it as the Rodriguez Structured Gaussian (RSG) method. Precisely, the RSG datasets with the following parameters were generated: The number of clusters \( k \in \{2, 10, 50\} \), the number of features \( d \in \{10, 50, 200\} \), and the number of objects per cluster \( N_c \in \{5, 50, 100\} \). It is worth noting that these parameter values were originally used in \cite{rodriguez2019clustering}. The RSG generator provides cluster-specific covariance matrices that are symmetric and positive semi-definite, ensuring realistic correlation structures between features. A cluster mixing parameter \( \alpha \) was tuned for each type of data to model a wide range of 
complexity levels and, as recommended in \cite{rodriguez2019clustering}, to ensure that no clustering algorithm would achieve an accuracy of 0\% or 100\%.

Fourth, to simulate high-dimensional anisotropic clusters with substantial inter-class separation, we employed the Repliclust cluster generation method recently proposed by Zellinger and Bühlmann \cite{zellinger2025natural}. Two cluster variants were considered: One with two clusters and one with five clusters. Each variant was instantiated in 10, 50, and 200 dimensions, with $n=2,000$ objects, and the number of objects per cluster $N_c = 1000$ for datasets with 2 clusters and $N_c = 400$ for datasets with 5 clusters. In both cases, the clusters were located along specific feature dimensions while maintaining centroid separation. Such a configuration provides a challenging benchmark for evaluating clustering robustness in high-dimensional settings.

\begin{figure}[H]
\centering
{\setlength{\fboxsep}{2pt}\fbox{\includegraphics[width=0.97\linewidth]{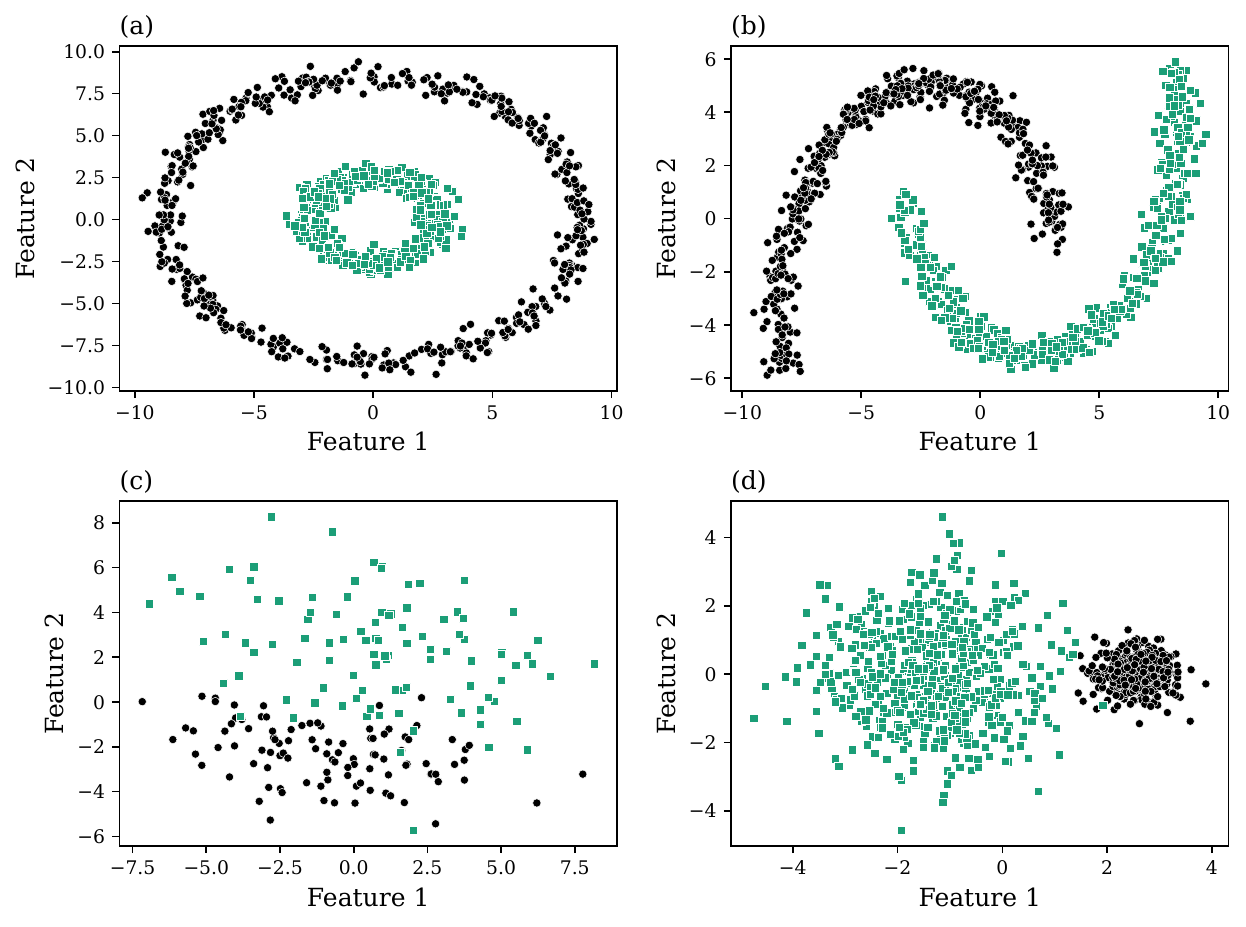}}}
\caption{Examples of synthetic datasets with two clusters and 50 original dimensions, 
    visualized by projecting raw data onto the first two principal components. The datasets correspond to the following cluster generation methods: (a) Circles, (b) Moons, (c) Rodriguez Structured Gaussian (RSG), and (d) Repliclust.}
\label{fig:exemple}
\end{figure}

To emulate real-world conditions in which some features are affected by random noise of different degree, we applied a structured noise injection scheme. All features were first normalized using z-scores. Then, we added Gaussian noise to 75\% of the original features. One-fourth of the features received standard Gaussian noise with (\(\mu = 0 \), \( \sigma = 1 \)), another one-fourth received Gaussian noise with (\(\mu = 0 \), \( \sigma = 0.5 \)), another one-fourth received Gaussian noise with (\(\mu = 0 \), \( \sigma = 0.25 \)), while the remaining quarter of the features remained unperturbed. This noise generation scheme, first used in \cite{makarenkov07error}, introduces heterogeneous noise distributions, enabling rigorous combined stress-testing of dimensionality reduction techniques and clustering algorithms.

For the Moons, Circles, and Repliclust data, we generated 50 different datasets per parameter ($N_c, d, k$) configuration considered (300 in total for each of them), while for the RSG data, we directly used the 265 synthetic datasets publicly released by Rodriguez et al. that span three cluster counts ($k \in {2, 10, 50}$), three dimensionalities ($d \in {10, 50, 200}$), and three numbers of objects per cluster ($N_c \in {5, 50, 100}$), as defined in \cite{rodriguez2019clustering}. 
In total, our experimental setup comprised 
1,165 different synthetic datasets, which are available in our GitHub repository at: \url{https://github.com/ELounissi/Dimensionality_Reduction_Clustering}. Figure \ref{fig:exemple} presents four typical examples of the Moons, Circles, RSG, and Repliclust datasets, visualized using their first two principal components.

\subsubsection{Real-world data} \label{subsubsec:real_world_data}

To complement the described synthetic benchmarks, we evaluated the methods' performances on 20 real-world datasets from the UCI repository\footnote{https://archive.ics.uci.edu}. These datasets, originally proposed as a standard benchmark for clustering algorithm comparison in the seminal work of Arbelaitz et al. \cite{arbelaitz2013extensive}, span a diverse range of domains, object, cluster, and feature numbers, as well as data distributions. This diversity ensures a comprehensive evaluation of clustering algorithms under realistic conditions, including varying levels of noise, nonlinearity, and class imbalance. The key characteristics of these datasets, including the number of objects, features, and clusters, are summarized in Table \ref{tab:uci_datasets}. They encompass application areas such as biology, medicine, chemistry, speech recognition, and image analysis. Each dataset is fully labeled, so external clustering validity indices can be computed from the ground-truth class memberships.

\begin{table}[H]
\centering
\caption{Real-world datasets from the UCI Machine Learning Repository used in our experiments (see also Arbelaitz et al. \cite{arbelaitz2013extensive}).}
\label{tab:uci_datasets}
    \begin{tabular}{lccc}
        \toprule
        \textbf{Datasets} & \textbf{Objects} & \textbf{Features} & \textbf{Clusters} \\
        \midrule
Breast tissue & 106 & 9 & 6 \\
Breast Wisconsin & 569 & 30 & 2 \\
Ecoli & 336 & 7 & 8 \\
Glass & 214 & 9 & 6 \\
Haberman & 306 & 3 & 2 \\
Ionosphere & 351 & 34 & 2 \\
Iris & 150 & 4 & 3 \\
Movement libras & 360 & 90 & 15 \\
Musk & 476 & 166 & 2 \\
Parkinsons & 195 & 22 & 2 \\
Segmentation & 2310 & 19 & 7 \\
Sonar all & 208 & 60 & 2 \\
Spectf & 267 & 44 & 2 \\
Transfusion & 748 & 4 & 2 \\
Vehicle & 846 & 18 & 4 \\
Vertebral column & 310 & 6 & 3 \\
Vowel context & 990 & 10 & 11 \\
Wine & 178 & 13 & 3 \\
Wine quality red & 1599 & 11 & 6 \\
Yeast & 1484 & 8 & 10 \\
\bottomrule
    \end{tabular}
\end{table}

\subsection{Dimensionality Reduction Methods} \label{subsec:dr_methods}

In our experiments, we considered five dimensionality reduction methods to project high-dimensional data into lower-dimensional spaces: Principal Component Analysis (PCA), Kernel Principal Component Analysis (Kernel PCA), Variational Autoencoder (VAE), Isometric Mapping (Isomap), and Multidimensional Scaling (MDS). These methods encompass both linear and nonlinear approaches, providing a comprehensive framework for evaluating how different dimensionality reduction methods influence clustering performance. In all our experiments, the \texttt{scikit-learn} implementations were used for PCA, Kernel PCA, Isomap, and MDS, and a PyTorch implementation was used for VAE. 
For all dimensionality reduction methods, we considered three target dimensionalities: $k-1$ dimensions, 25\% of the original number of dimensions, and 50\% of the original number of dimensions. For the $k-1$ setting, we used $\max(k-1,2)$ to avoid reduction to a single dimension.

\subsubsection{Principal Component Analysis (PCA)}
PCA is a widely used linear dimensionality reduction technique that projects data onto a lower-dimensional orthogonal subspace by identifying directions (known as principal components) that maximize variance \cite{pearson1901liii}. Given a dataset with $n$ objects and \(d\) features, PCA computes the covariance matrix of the centered data and derives its eigenvectors. The top \(m\) eigenvectors, corresponding to the largest eigenvalues, form the new basis. The data is then projected onto this \(m\)-dimensional subspace. PCA is effective when the data lies in a low-dimensional linear subspace, serving as a natural comparative baseline. Except \textit{n\_components} and \textit{svd\_solver}='full', all other PCA settings were its default \texttt{scikit-learn} settings.

\subsubsection{Kernel Principal Component Analysis (Kernel PCA)}
Kernel PCA extends PCA to nonlinear settings by implicitly mapping the data into a high-dimensional feature space via kernel functions and then performing PCA in this transformed space \cite{scholkopf1998kpca}. A kernel matrix \(K\) is constructed using pairwise similarities, commonly with a radial basis function:
\begin{equation}
K_{ij} = \exp\left(-\gamma \, \|x_i - x_j\|^2\right),
\end{equation}
where \(\gamma > 0\) controls the kernel scale (inverse squared bandwidth), and $x_i$ and $x_j$ are two given objects. The centered kernel matrix is then decomposed to extract principal components in the kernel-induced space. This approach enables Kernel PCA to uncover nonlinear structures, although its scalability is limited by the need to compute the full \(n \times n\) kernel matrix. Except \textit{n\_components}, \textit{kernel}='rbf', \textit{eigen\_solver}='arpack', $\gamma=1/d$, and \textit{random\_state} set to the current random seed, all other Kernel PCA settings were the default \texttt{scikit-learn} settings. It is worth noting that the experiments conducted with the 'linear', 'poly', 'rbf', and 'sigmoid' kernel functions showed that the best average clustering results were obtained using the 'rbf' kernel. 

\subsubsection{Variational Autoencoder (VAE)}

VAEs are generative models that learn probabilistic latent representations through neural networks \cite{kingma2013vae}. Unlike deterministic autoencoders, VAEs approximate the posterior distribution over latent variables using a variational inference framework. Given an observed variable \(x\) and a latent variable \(z\), the marginal log-likelihood can be defined as follows:
\begin{equation}
\log p_\theta(x) = \log \int p_\theta(x|z)\, p(z)\, dz,\end{equation}
\noindent and then approximated using the evidence lower bound (ELBO):
\begin{equation}
\mathcal{L}(\theta,\phi; x) = \mathbb{E}_{q_\phi(z|x)} [\log p_\theta(x|z)] 
- D_{\mathrm{KL}}\!\big(q_\phi(z|x)\,\|\,p(z)\big),
\end{equation}
where \(q_\phi(z|x)\) is the encoder, \(p_\theta(x|z)\) is the decoder, and \( D_{KL}(\cdot || \cdot) \) denotes the Kullback–Leibler divergence that regularizes the latent distribution toward the prior \(p(z)\), typically a standard normal distribution \( \mathcal{N}(0, I) \). VAEs are well-suited for capturing complex, nonlinear manifolds in high-dimensional data.

In this study, we employed a symmetric encoder–decoder design. The encoder has two fully connected layers (with 64 and 32 neurons, and ReLU), followed by Batch Normalization and Dropout (0.4). The latent space is parameterized by \(\mu_z\) and \(\log \sigma^2_z\) of size \(n_{\text{components}}\), with sampling:
\begin{equation}
z = \mu_z + \exp\!\big(0.5\, \log \sigma^2_z\big) \odot \epsilon.
\end{equation}
The decoder mirrors the encoder (with 32 and 64 neurons, ReLU, BatchNorm, and Dropout \(=0.4\)) and outputs a \(d\)-dimensional vector 
through a linear output layer. Batch Normalization and Dropout accelerate convergence, improve generalization, and mitigate posterior collapse. We trained the VAE using the Adam optimizer (learning rate of 0.001 and gradient norm clipping at 10), squared-error reconstruction loss, and a KL-divergence term with a weight of 1 and a batch size of 64 for 100 epochs on all objects of each dataset. After training, we retained \(z_{\text{mean}}\) as a deterministic embedding for clustering. This VAE configuration provided the best average clustering results among several VAE configurations tested in our experiments.

\subsubsection{Isometric Mapping (Isomap)}

Isomap is a nonlinear dimensionality reduction technique based on manifold learning that preserves geodesic distances rather than Euclidean distances \cite{tenenbaum2000isomap}. Given a set of objects $X = \{x_i\}_{i=1}^n \subset \mathbb{R}^d$, a $k$-nearest-neighbor graph is constructed, where edges are weighted by Euclidean distances. Geodesic distances, $D_M$, are then estimated as shortest-path distances in this graph. Classical multidimensional scaling (cMDS) is subsequently applied to $D_M$ to calculate the double-centered Gram matrix:
\begin{equation}
    G = -\tfrac{1}{2}\, H (D_M \odot D_M) H,
\end{equation}
where $H = I - \tfrac{1}{n}\mathbf{1_n}\mathbf{1_n^\top}$, extracting the top $m$ eigenpairs. The resulting embedding minimizes the discrepancy between the geodesic and Euclidean distances in a low-dimensional space. Isomap was implemented using \texttt{scikit-learn}. Its number of neighbors setting was set to 5 and doubled until the neighborhood graph became connected.

\subsubsection{Multidimensional Scaling (MDS)}
MDS seeks a low-dimensional embedding in which Euclidean distances approximate a given dissimilarity matrix \cite{kruskal1964multidimensional}. This makes MDS suitable when similarities are defined by arbitrary metrics, such as correlations or domain-specific measures. Given dissimilarities $\{\delta_{ij}\}$, MDS finds a configuration $Z$ that minimizes Kruskal’s stress function:
 
 \begin{equation}
\mathrm{Stress}(Z) \;=\; \sqrt{\frac{\sum_{i<j} \big(\delta_{ij} - \delta'_{ij}(Z)\big)^2}{\sum_{i<j} \delta_{ij}^2}},
\end{equation}
where $\delta'_{ij}(Z)$ denotes Euclidean distances in the embedding. In our experiments, we used the \texttt{scikit-learn} implementation of MDS that optimizes this criterion via iterative majorization with a PCA-initialized start. Except \textit{n\_components}, \textit{random\_state=10}, \textit{max\_iter=300}, and \textit{eps=1e-6}, all other MDS settings were its default \texttt{scikit-learn} settings.

\subsection{Clustering Algorithms} \label{subsec:clust_alg}

In this study, we considered four popular clustering algorithms, each representing a distinct clustering paradigm, including centroid-based, probabilistic, hierarchical, and density-based clustering.

\subsubsection{$K$-means} $K$-means is a centroid-based clustering algorithm that partitions \(n\) data points (i.e., objects) into \(k\) non-overlapping clusters by minimizing the within-cluster sum of squared Euclidean distances. It iteratively assigns points to the nearest cluster centroid and updates centroids as the means of assigned points until the convergence is reached \cite{macqueen1967some}. While very fast, $k$-means assumes Gaussian clusters of equal variance and is sensitive to initialization and outliers \cite{de2023k,de2016applying,PENA19991027,de2026improving}. To improve its robustness against random initializations and reduce variance across runs, we adopted the \textit{$k$-means++} initialization strategy, repeating clustering 100 times ($n\_init = 100$) with different starting random partitions.
\subsubsection{Agglomerative Hierarchical Clustering (AHC)} 
AHC is a bottom-up clustering algorithm that initially treats each object as an individual cluster and successively merges the closest object pairs until a stopping condition (e.g., a pre-defined number of clusters \(k\)) is met \cite{sokal1963numerical}. In our experiments, we considered different affinity measures (\textit{euclidean}, $l1$, $l2$, \textit{manhattan}, and \textit{cosine}) along with multiple linkage strategies (\textit{complete}, \textit{average}, \textit{single}, and \textit{ward}), which allowed us to systematically assess the influence of distance measure and merging algorithm selection on clustering outcomes. Note that the \textit{ward} linkage strategy is restricted to Euclidean distances. 
For each dataset, we evaluated all valid combinations of affinity measure and linkage strategy on the unreduced data and on each reduced representation. In each case, we selected the clustering solution with the highest Silhouette score.

\subsubsection{Gaussian Mixture Model (GMM)} GMM is a probabilistic clustering model that assumes that data is generated from a mixture of Gaussian distributions with unknown parameters \cite{wolfe1970pattern}. Each object is assigned a probability of belonging to each GMM component (i.e., cluster), and parameters (means, covariances, and 
mixing weights) are estimated via the Expectation–Maximization (EM) algorithm, enabling soft assignments and flexible modeling of elliptical cluster shapes. In our experiments, we evaluated GMM under different covariance parameterizations (\textit{spherical}, \textit{tied}, \textit{diag}, and \textit{full}) that determine the shape and flexibility of the Gaussian components. 
For each dataset, we fitted models with all the four covariance parameterizations on the unreduced data and on each reduced representation, using 10 initializations and a covariance regularization of $10^{-5}$. In each case, we selected the model providing the lowest value of Bayesian Information Criterion (BIC).

\subsubsection{Ordering Points To Identify the Clustering Structure (OPTICS)} OPTICS is a density-based clustering algorithm that constructs an augmented ordering of the given set of objects to retrieve clusters of varying shapes and densities \cite{ankerst1999optics}. Unlike DBSCAN \cite{ester1996density}, it does not require a single global density threshold. OPTICS identifies reachability and core distances to construct a reachability plot from which clusters can subsequently be extracted without requiring a single global density threshold. In our experiments, we varied its key parameters to examine their impact on cluster detection: \textit{min\_samples} ranged from 5 to 10 (with a step of 1), the cluster extraction method was set to \textit{xi} (with $\xi = 0.05$), and for the \textit{min\_cluster\_size} parameter, we tested minimum cluster sizes ranging from 0\% to 50\% of the dataset size with increments of 5\% (i.e., 0\%, 5\%, 10\%, \ldots, 50\%).

For each dataset, we evaluated all combinations of these parameter values on the unreduced data and on each reduced representation. In each case, we selected the clustering solution whose number of clusters was the closest to $k$. Ties were resolved by selecting the solution with the smallest fraction of noise objects. Each noise object was then assigned to the cluster with the nearest centroid.

\subsection{Evaluation Metric} \label{subsec:ari}
We used the Adjusted Rand Index (ARI) \cite{hubert1985comparing} to quantify the agreement between an obtained clustering solution and ground-truth labels, which were available for all synthetic and real-world datasets considered in our study. ARI is a chance-adjusted variant of the Rand Index varying in the range $[-1,1]$; random, independent partitions have an expected ARI close to $0$, perfect agreement yields $1$, while negative ARI values indicate worse-than-random agreement.

Consider a dataset of $n$ items with two partitions $X=\{X_1,\dots,X_r\}$ and $Y=\{Y_1,\dots,Y_s\}$. Let $n_{ij}=|X_i \cap Y_j|$ denote the entries of the contingency table, with row and column sums $a_i=\sum_{j=1}^{s} n_{ij}$ and $b_j=\sum_{i=1}^{r} n_{ij}$, respectively. ARI is then defined as follows:
\begin{equation}
\label{eq:ari}
\mathrm{ARI} =
\frac{\sum_{i=1}^{r}\sum_{j=1}^{s} \binom{n_{ij}}{2}
- \left[ \sum_{i=1}^{r} \binom{a_i}{2} \sum_{j=1}^{s} \binom{b_j}{2} \right] \! / \binom{n}{2}}
{\frac{1}{2}\!\left[ \sum_{i=1}^{r} \binom{a_i}{2} + \sum_{j=1}^{s} \binom{b_j}{2} \right]
- \left[ \sum_{i=1}^{r} \binom{a_i}{2} \sum_{j=1}^{s} \binom{b_j}{2} \right] \! / \binom{n}{2}}.
\end{equation}
Here, $\binom{m}{2}$ denotes the number of unordered pairs from $m$ items.

\subsection{Experimental Pipeline}\label{subsec:exp_setup}

Our experimental pipeline includes four main stages. All experiments were carried out using Python 3.12 along with \texttt{scikit-learn} 1.7.1 and PyTorch 2.9.1 libraries.

First, all datasets were preprocessed using z-score normalization, ensuring that each feature has zero mean and unit variance. This step mitigates the influence of scale differences across dimensions, which is crucial for distance-based clustering and projection techniques. 

Second, clustering was performed directly on the normalized unreduced datasets using four representative algorithms: $k$-means, AHC, GMM, and OPTICS. The quality of the resulting cluster partitions was evaluated using ARI. For the 20 real-world datasets, every clustering run was repeated with 10 different random seeds and the ARI scores were averaged per dataset, while each of the 1,165 synthetic datasets was processed with a single seed.

Third, five dimensionality reduction techniques (PCA, Kernel PCA, VAE, Isomap, and MDS) were applied to each normalized unreduced dataset to project it into a lower-dimensional subspace. We considered three levels of dimensionality reduction recommended in the literature:
(1) Compression to $k-1$ dimensions, where $k$ is the number of clusters, which is motivated by the subspace bound. For example, Ding et al. \cite{ding2002adaptive} argued that the effective clustering dimensions for $k$ spherical Gaussians are spanned by $k$ centers for which the dimensionality of the relevant clustering subspace equals $k-1$;
(2) Another drastic dimensionality compression considered is to 25\% of the original number of features \cite{tang2005comparing, karypis2000concept}; 
(3) A moderate natural compression to 50\% of the original number of features \cite{hasan2021review}.

These settings enable evaluation of clustering performance under both theoretically motivated and practically relevant dimensionality constraints. Clustering algorithms were then re-applied to the lower-dimensional data, and the ARI scores were computed to assess performance changes.

Fourth, clustering outcomes with and without dimensionality reduction were compared to analyze the impact of each dimensionality reduction method on clustering performance across various dataset geometries and dimensionalities. Such a comparison helped us to discover the strengths and limitations of different dimensionality reduction method/clustering algorithm pairings across various data types.

\begin{figure}[H]
\centering
{\setlength{\fboxsep}{2pt}\fbox{\includegraphics[width=0.97\linewidth]{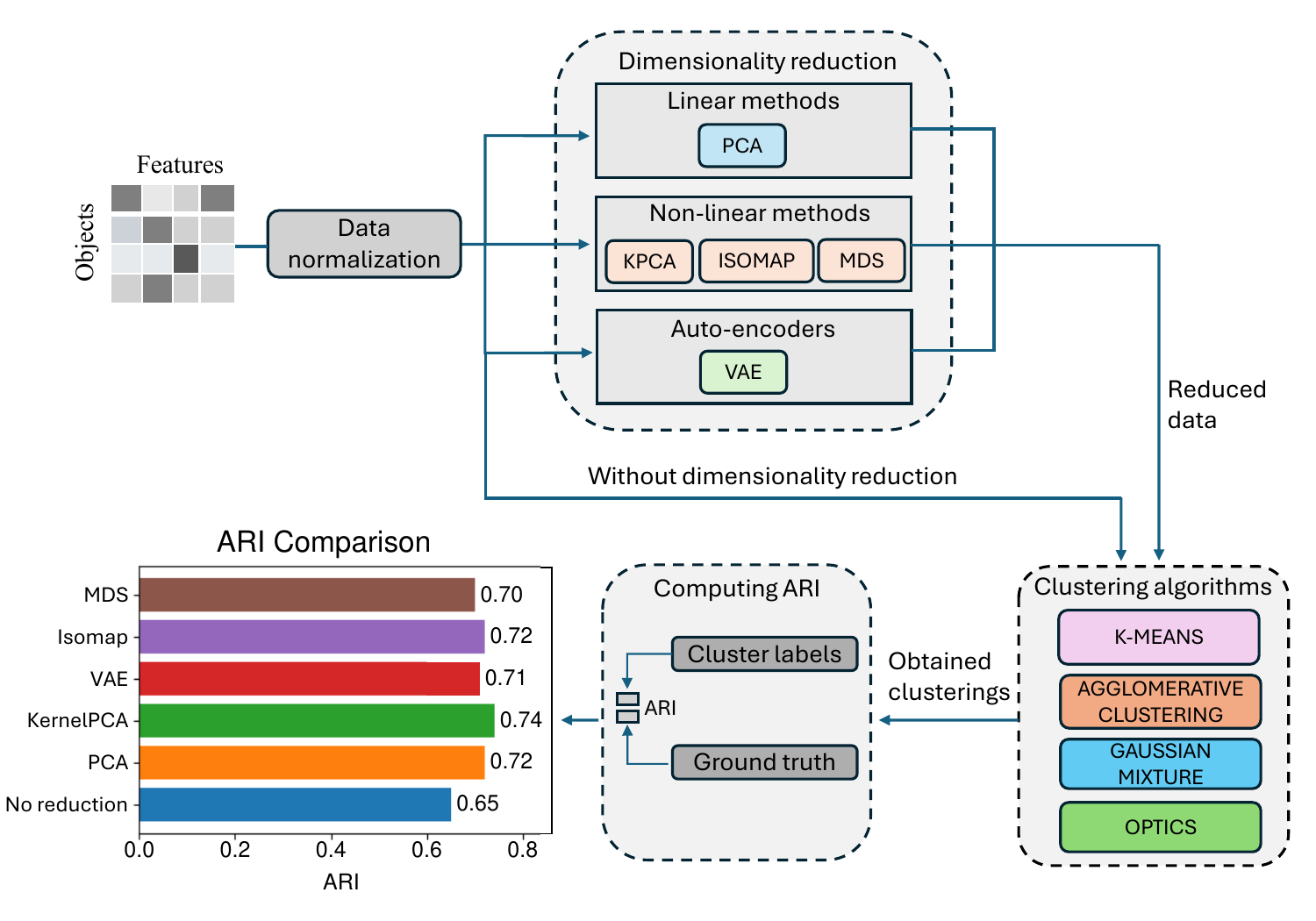}}}
\caption{Overview of the evaluation pipeline assessing the impact of dimensionality reduction on clustering.}
\label{fig:system_design}
\end{figure}

Figure \ref{fig:system_design} summarizes our experimental pipeline, from data preprocessing through dimensionality reduction, clustering, and ARI-based evaluation.

\section{Results}\label{sec:results}

\subsection{Experiments with Synthetic Data} \label{subsec:results_synth_data}

\begin{figure}[H]
\centering
{\setlength{\fboxsep}{2pt}\fbox{\includegraphics[width=0.97\linewidth]{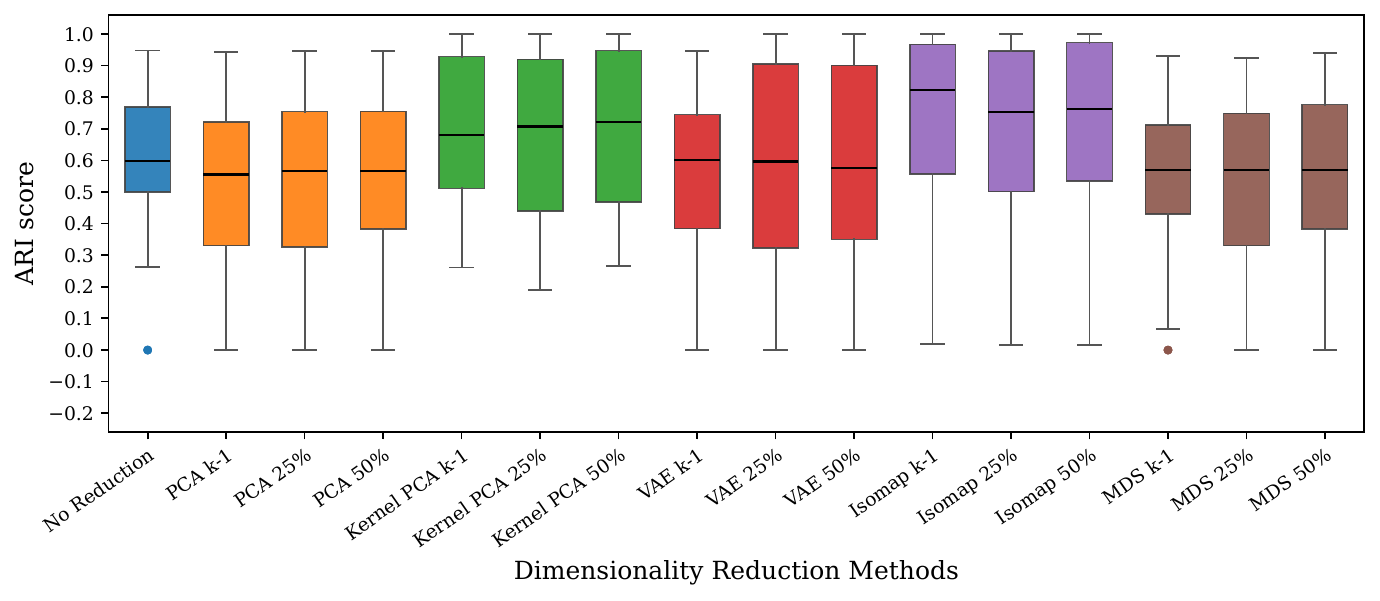}}}
\caption{Boxplot summarizing ARI scores for different dimensionality reduction methods applied to synthetic datasets, when they were followed by clustering with $k$-means.}
\label{fig:k_means_synthetic}
\end{figure}

\begin{figure}[H]
\centering
{\setlength{\fboxsep}{2pt}\fbox{\includegraphics[width=0.97\linewidth]{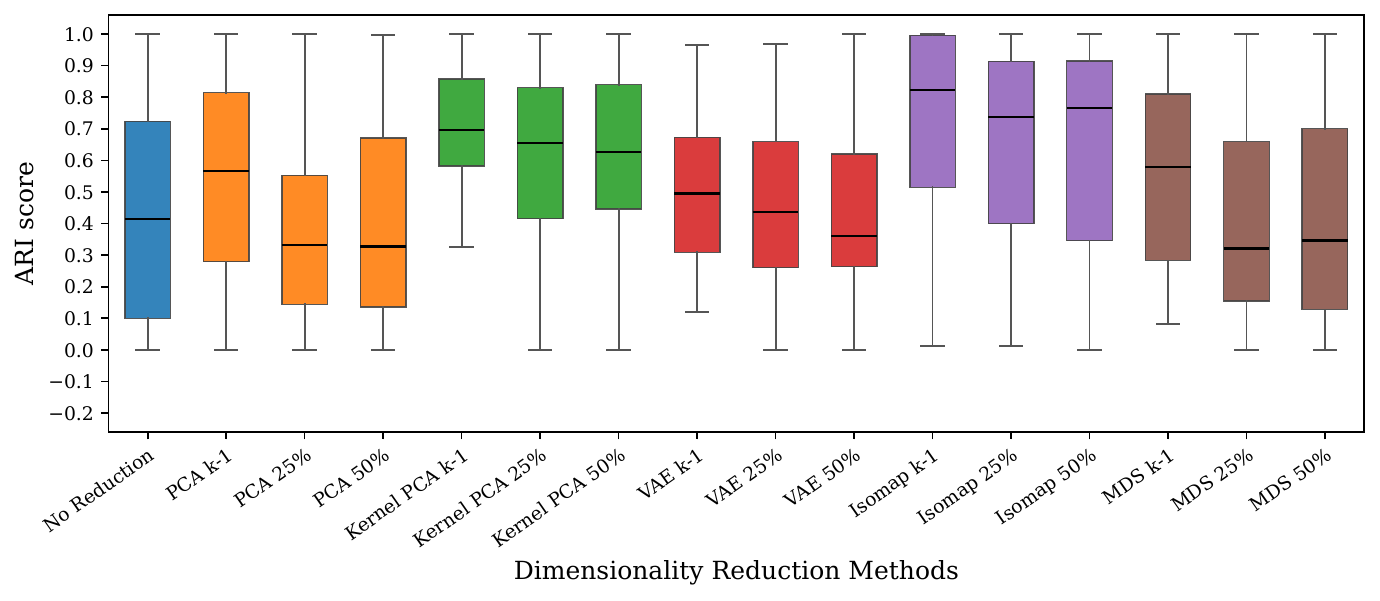}}}
\caption{Boxplot summarizing ARI scores for different dimensionality reduction methods applied to synthetic datasets, when they were followed by Agglomerative Hierarchical Clustering (AHC).}
\label{fig:agglo_synthetic}
\end{figure}

\begin{figure}[H]
\centering
{\setlength{\fboxsep}{2pt}\fbox{\includegraphics[width=0.97\linewidth]{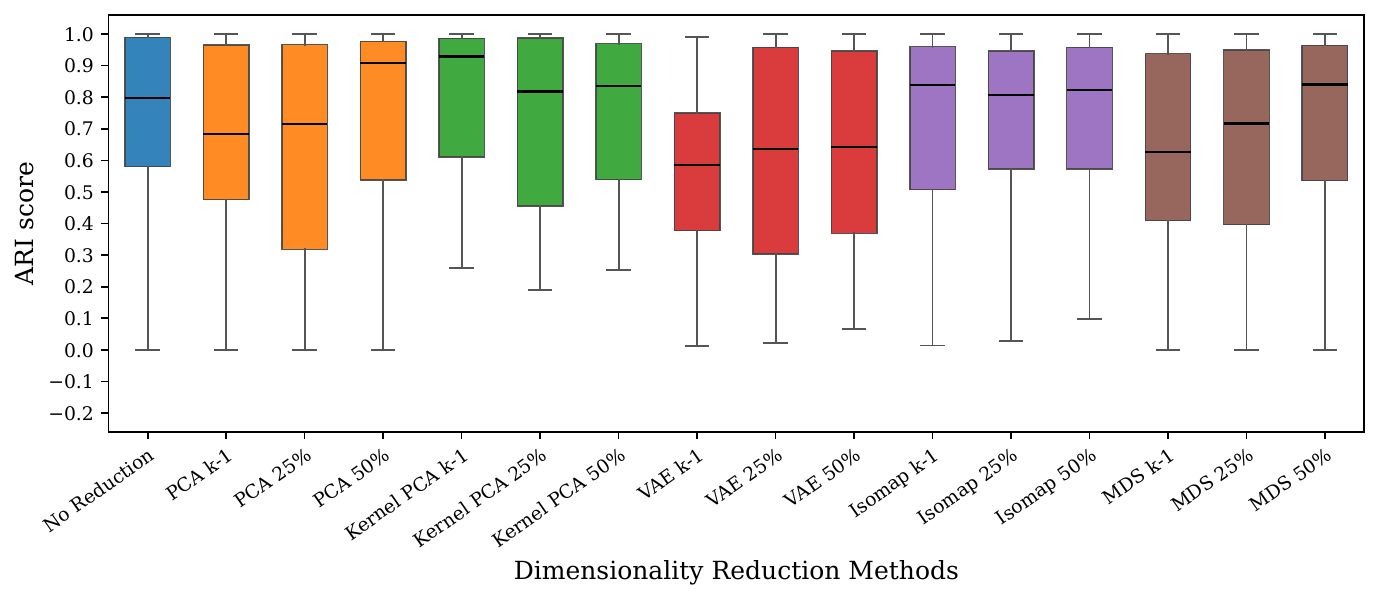}}}
\caption{Boxplot summarizing ARI scores for different dimensionality reduction methods applied to synthetic datasets, when they were followed by clustering with Gaussian Mixture Models (GMM).
    }
\label{fig:gaussian_synthetic}
\end{figure}

\begin{figure}[H]
\centering
{\setlength{\fboxsep}{2pt}\fbox{\includegraphics[width=0.97\linewidth]{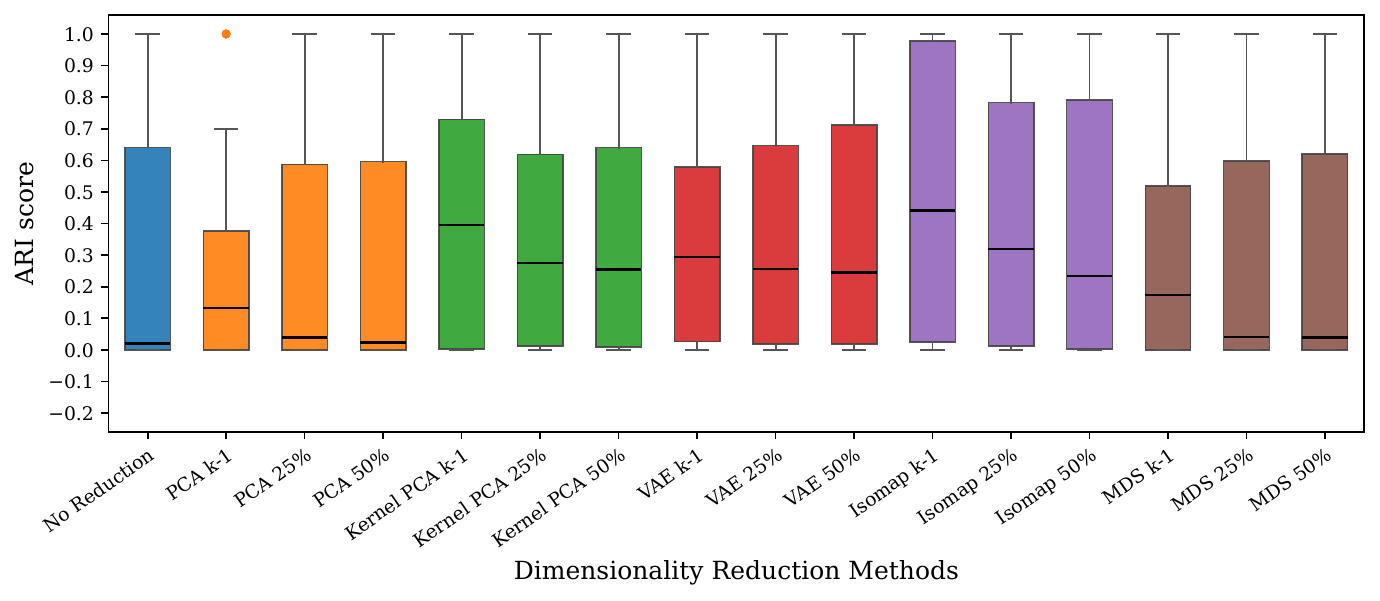}}}
\caption{Boxplot summarizing ARI scores for different dimensionality reduction methods applied to synthetic datasets, when they were followed by clustering with OPTICS.}
\label{fig:optic_synthetic}
\end{figure}

Boxplot diagrams in Figures \ref{fig:k_means_synthetic} to \ref{fig:optic_synthetic} summarize ARI-based dimensionality reduction performances per clustering algorithm, while Tables \ref{tab:ari_circles} to \ref{tab:ari_repliclust} in \ref{sec:apx_a} report the corresponding average ARI scores per data type. 

As we can observe on the boxplots, across the four clustering algorithms considered, applying a nonlinear dimensionality reduction often improves ARI relative to no reduction, though the observed effect is data type- and clustering algorithm-dependent. 
Kernel PCA is very competitive when used with $k$-means and OPTICS, especially at the 50\% dimensionality reduction level for $k$-means and at the $k-1$ reduction level for OPTICS.
Isomap often yields the best results when paired with AHC and OPTICS, while its best improvements are observed at the $k-1$ dimensionality reduction level. MDS shows a data type sensitive behavior: While it mostly underperforms or remains close to no-reduction baseline on Circles, Moons, and RSG data, it provides notable gains on Repliclust data when paired with AHC, but clear losses at the $k-1$ reduction level when paired with $k$-means and GMM (see Table \ref{tab:ari_repliclust}). VAE's performance depends on both the data type and the clustering algorithm. It yields lower mean ARI than no-reduction baseline on RSG data with all the four clustering algorithms, but improves the baseline results with OPTICS on Circles and Moons, and with both AHC and OPTICS on Repliclust data. Detailed win rates and average win/loss ARI changes relative to no-reduction baseline are reported in Section \ref{subsec:agg_analysis} (see Tables \ref{tab:kmeans_aggregate} to \ref{tab:optics_aggregate}).

Figure \ref{fig:k_means_synthetic} summarizes the ARI distributions over different data types for $k$-means clustering and all dimensionality reduction methods and levels considered. Here, Kernel PCA provides strong overall improvements, especially at the 50\% dimensionality reduction level, while Isomap has the highest median at the $k-1$ reduction level. Precisely, Kernel PCA achieves the best ARI performance on Circles and Repliclust data (see Tables \ref{tab:ari_circles} and \ref{tab:ari_repliclust}), while Isomap remains superior to Kernel PCA on Moons and RSG data (see Tables \ref{tab:ari_moons} and \ref{tab:ari_rodriguez}). MDS remains close to no-reduction baseline on Circles and Moons data, but underperforms on Repliclust data, particularly at the $k-1$ dimensionality reduction level. The VAE results also stay close to baseline, with slight gains on Moons and Repliclust data, but lower ARI scores on RSG data. Importantly, no-reduction baseline was rarely best-performing, underscoring that a suitable nonlinear reduction can substantially improve cluster separability for centroid-based clustering algorithms.

Figure \ref{fig:agglo_synthetic} illustrates the ARI distributions for Agglomerative Hierarchical Clustering (AHC) and all dimensionality reduction methods and levels considered. Nonlinear manifold-based approaches prove most effective here, with Isomap delivering the highest median ARI values, particularly at the $k-1$ dimensionality reduction level. This effect is most pronounced on Moons and RSG data (see Tables \ref{tab:ari_moons} and \ref{tab:ari_rodriguez}), where Isomap consistently surpasses both PCA and Kernel PCA. Kernel PCA also maintains competitive performance across datasets, and it achieves the best ARI performance on Circles and Repliclust data (see Tables \ref{tab:ari_circles} and \ref{tab:ari_repliclust}), where Isomap is much weaker. Linear PCA provides a clear gain only at the $k-1$ reduction level, mainly on Repliclust data, and remains below Kernel PCA for all the four data types. MDS exhibits strong data-dependent variability: While ineffective on Circles, it provides meaningful gains on RSG (at the $k-1$ reduction level) and Repliclust data. VAE provides moderate gains for AHC, mainly on Repliclust data, with its highest median at the $k-1$ reduction level, but it underperforms on RSG data. Overall, no-reduction baseline was rarely best-performing, confirming that nonlinear reduction, particularly that performed with Isomap and Kernel PCA, substantially enhances clustering performance of hierarchical agglomerative clustering algorithms.

Figure \ref{fig:gaussian_synthetic} presents the ARI results for GMM across all dimensionality reduction methods and levels considered. The advantage of nonlinear reductions is less pronounced for mixture models and depends strongly on the data type. Isomap shows a substantial improvement on Moons data, particularly at the $k-1$ dimensionality reduction level (see Table \ref{tab:ari_moons}), but performs poorly on Repliclust data. Kernel PCA maintains strong and stable performance, achieving the highest median at the $k-1$ reduction level and providing the best ARI performance on Circles and Repliclust data. Linear PCA remains competitive, with the best ARI performance on RSG data at the 50\% reduction level (see Table \ref{tab:ari_rodriguez}), but is outperformed by nonlinear methods on the curved Circles and Moons manifolds. VAE yields the lowest median ARI among the dimensionality reduction methods at each reduction level. It improves mean ARI over no-reduction baseline only on Circles, while reducing it on Moons, RSG, and Repliclust data (see Table \ref{tab:ari_rodriguez}). MDS remains data-sensitive, with strong results on Repliclust data at the 25\% and 50\% reduction levels, but weaker performance at the $k-1$ reduction level (see Table \ref{tab:ari_repliclust}), and much weaker results on Circles data (see Table \ref{tab:ari_circles}). Importantly, dimensionality reduction performed with Kernel PCA and Isomap can still enhance clustering performance of Gaussian Mixture Models, but the gains are smaller and less consistent than for AHC, and none of them is statistically significant (see Table \ref{tab:Wilcoxon_synthetic}).

Figure \ref{fig:optic_synthetic} depicts the ARI performances for OPTICS across all dimensionality reduction methods and  levels considered. The OPTICS-related results are highly data type-dependent, with strong general improvements over no-reduction baseline on Circles, Moons, and Repliclust data, but markedly poor outcomes on RSG data, where only Isomap improves over baseline. Kernel PCA and Isomap achieve the highest medians at the $k-1$ reduction level. Kernel PCA also provides the best ARI performance on Circles (see Table \ref{tab:ari_circles}) and Repliclust data (see Table \ref{tab:ari_repliclust}), where its reduction to $k-1$ dimensions raises the ARI from 0.002 to 0.688. Isomap, consistent with its success with AHC, improves over no-reduction baseline on all the four data types, providing the best ARI performance on Moons and RSG data (see Tables \ref{tab:ari_moons} and \ref{tab:ari_rodriguez}). Linear PCA displays variable behavior, with gains at the $k-1$ reduction level on Circles, Moons, and Repliclust data, but notable ARI drops on RSG data. VAE improves mean ARI results over no-reduction baseline on Circles, Moons, and Repliclust data at all the three reduction levels, but yields lower mean ARI scores on RSG data. MDS provides clear gains only at the $k-1$ reduction level, notably on Moons and Repliclust data. No-reduction baseline is occasionally competitive but is rarely best-performing overall. These findings underscore that for density-based clustering, the benefits of dimensionality reduction are strongly data type-dependent, with no single dimensionality reduction method dominating across all experimental conditions.

\subsection{Experiments with Real-World Data}

\begin{figure}[H]
\centering
{\setlength{\fboxsep}{2pt}\fbox{\includegraphics[width=0.97\linewidth]{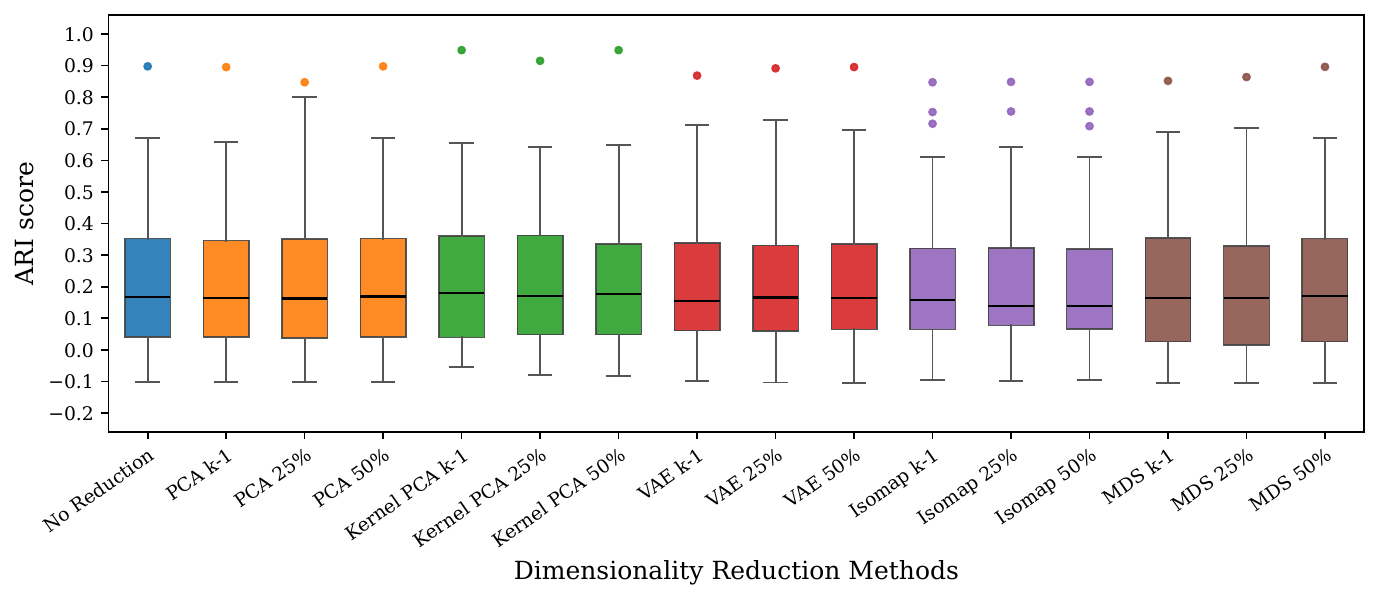}}}
\caption{Boxplot summarizing ARI scores for different dimensionality reduction methods applied to real-world datasets, when they were followed by clustering with $k$-means.}
\label{fig:k-means_real}
\end{figure}

Across the 20 real-world datasets from Table \ref{tab:uci_datasets}, dimensionality reduction methods only marginally improve the results of $k$-means obtained with unreduced data, with no single method consistently dominating. The corresponding boxplot diagram (see Fig. \ref{fig:k-means_real}) shows medians and spreads similar to no-reduction baseline, with a few high outliers for all methods (in particular for Isomap). Kernel PCA, Isomap, and VAE provide small average gains over no-reduction baseline at all the three dimensionality reduction levels (see Table \ref{tab:kmeans_aggregate}), but none of these improvements is statistically significant (see Table \ref{tab:Wilcoxon}). 
Overall, the results obtained using $k$-means on real-world data suggest that while dimensionality reduction can occasionally yield meaningful ARI improvements (e.g., on Iris, PCA used with 25\% dimensionality reduction substantially improved over baseline, providing an ARI gain of 0.18), baseline clustering in the original space remains a robust default option.

\begin{figure}[H]
\centering
{\setlength{\fboxsep}{2pt}\fbox{\includegraphics[width=0.97\linewidth]{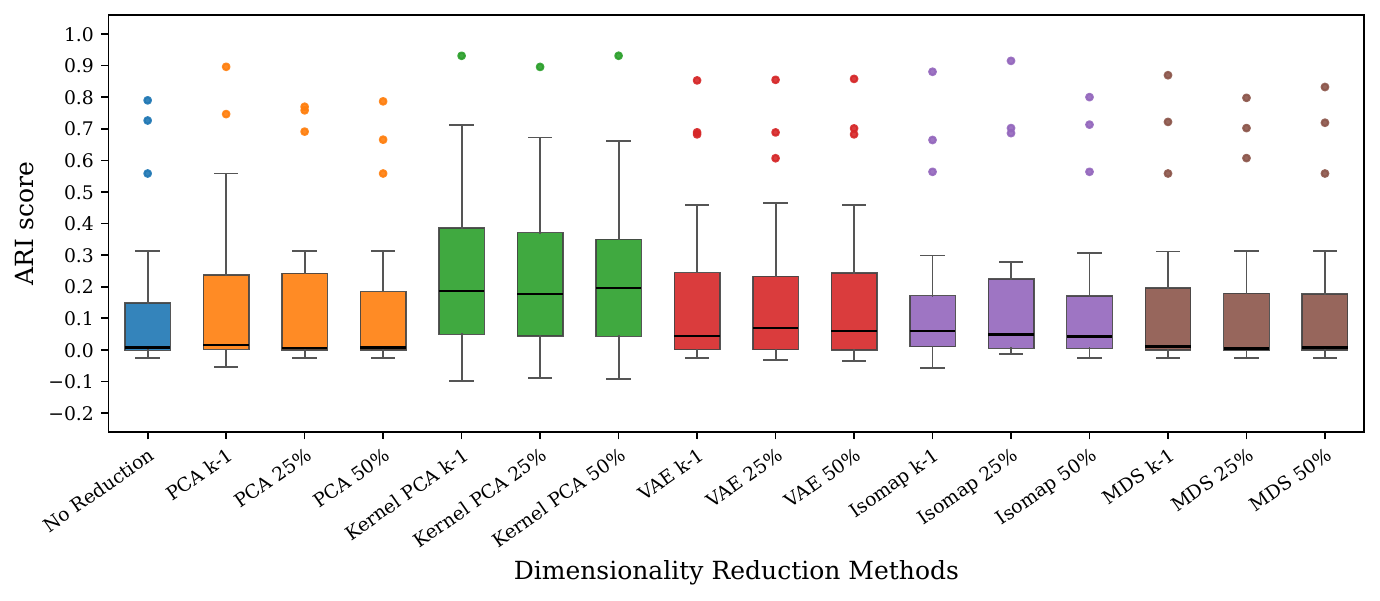}}}
\caption{Boxplot summarizing ARI scores for different dimensionality reduction methods applied to real-world datasets, when they were followed by Agglomerative Hierarchical Clustering (AHC).}
\label{fig:agglo_real}
\end{figure}

For hierarchical clustering, Kernel PCA provides the most visible improvements relative to no-reduction baseline (see Fig. \ref{fig:agglo_real}). When applied at the $k-1$ and 50\% dimensionality reduction levels, it achieves the most important gains compared to baseline 
(for detailed results, see also Table \ref{tab:agglomerative_real}). VAE and Isomap also provide smaller but statistically significant gains over baseline (see Table \ref{tab:Wilcoxon}), while PCA and MDS remain close to baseline, with sporadic gains, but no systematic advantage. 
Overall, AHC benefits most from Kernel PCA, while manifold and deep generative reduction methods (i.e., Isomap and VAE) provide more modest but consistent gains.

\begin{figure}[H]
\centering
{\setlength{\fboxsep}{2pt}\fbox{\includegraphics[width=0.97\linewidth]{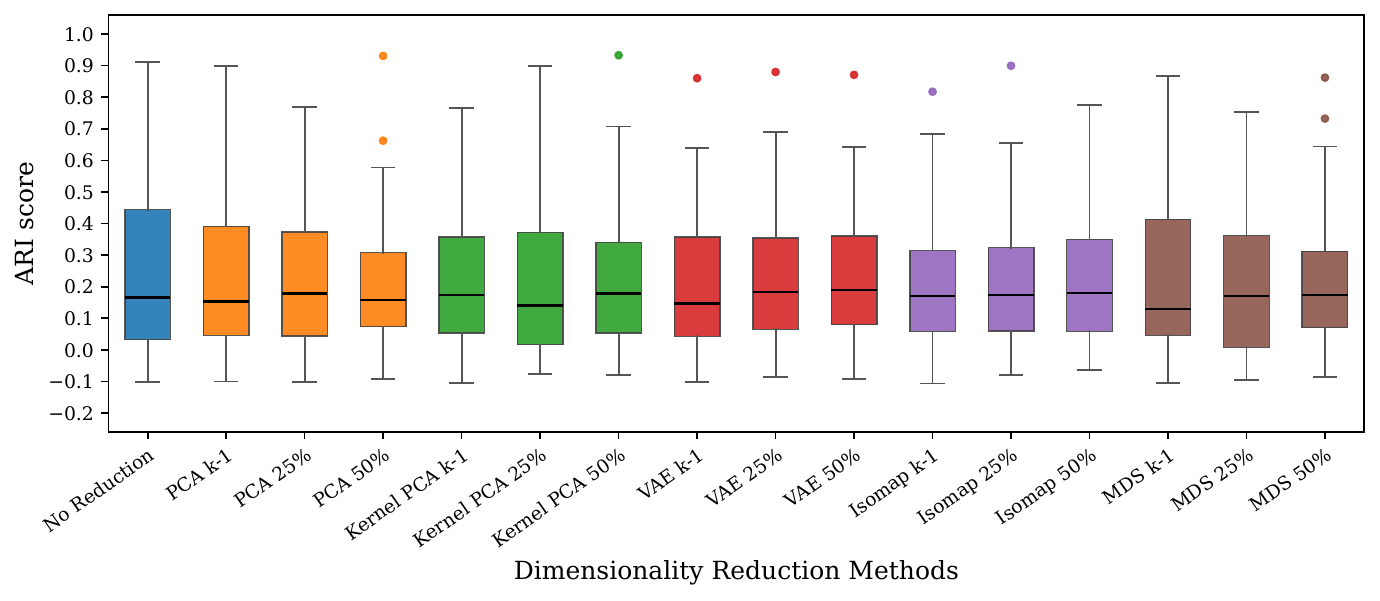}}}
\caption{Boxplot summarizing ARI scores for different dimensionality reduction methods applied to real-world datasets, when they were followed by clustering with Gaussian Mixture Models (GMM).}
\label{fig:gaussian_real}
\end{figure}

Figure \ref{fig:gaussian_real} summarizes the ARI distributions over 20 real-world datasets obtained using GMM. No-reduction baseline provides the highest average ARI here, as GMM in the original space performs best on several datasets (e.g., Breast Wisconsin, Ionosphere, Iris, and Vertebral column), where all dimensionality reduction methods lead to substantial losses. Kernel PCA produces a few of the highest single-dataset scores, notably on Breast tissue and Wine (the latter with ARI = 0.93 at the 50\% dimensionality reduction level), but its average performance at all the three reduction levels remains below that of no-reduction baseline (see Table \ref{tab:gaussian_real}), reflecting sporadic gains rather than a systematic advantage. Linear PCA occasionally matches or slightly exceeds baseline. MDS and VAE underperform on average as well, exhibiting a broad results dispersion in these settings. Overall, no dimensionality reduction method stands out as an aggregate winner when GMM clustering is used, all methods yield dataset-specific improvements only, and no-reduction baseline remains the best option on average.

\begin{figure}[H]
\centering
{\setlength{\fboxsep}{2pt}\fbox{\includegraphics[width=0.97\linewidth]{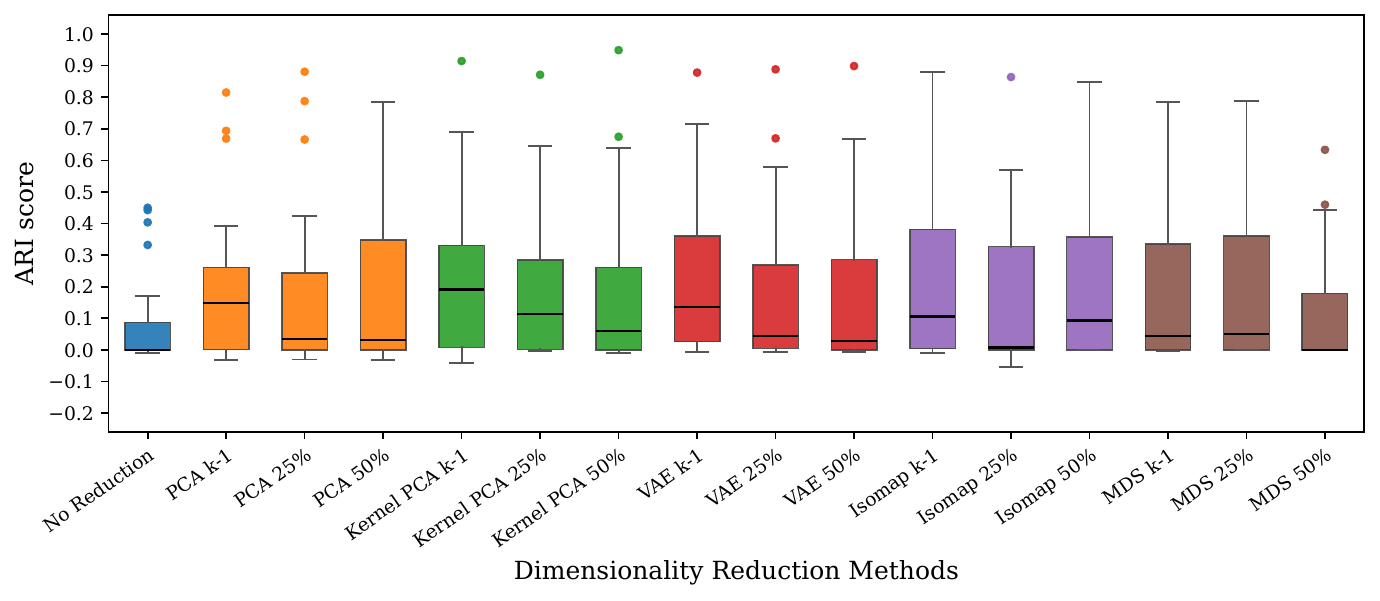}}}
\caption{Boxplot summarizing ARI scores for different dimensionality reduction methods applied to real-world datasets, when they were followed by clustering with OPTICS.}
\label{fig:optics_real}
\end{figure}

Figure \ref{fig:optics_real} presents the ARI distributions for OPTICS clustering carried out on 20 real-world datasets. The performance varies considerably across methods and datasets. Most ARI distributions can be characterized by low medians and large variability, but specific combinations provide substantial improvements relative to baseline. On aggregate, Kernel PCA, especially at the $k-1$ dimensionality reduction level, achieves the strongest improvements over no-reduction baseline, producing a much higher median and frequent high outliers, e.g., with datasets such as Breast Wisconsin, Ecoli, and Wine, where clear ARI improvements can be observed (see Table \ref{tab:optics_real}). Linear PCA also improves over no reduction baseline on average, with a clearly higher median at the $k-1$ reduction level, but is typically baseline-level at the two other reduction levels. Isomap and MDS improve on average as well, although their gains are less frequent and come from a few datasets (e.g., see the results for the Wine dataset in Table \ref{tab:optics_real}). VAE at the $k-1$ reduction level provides the second largest average gain after Kernel PCA, along with one of the highest medians. Overall, Kernel PCA and VAE with reduction to $k-1$ dimensions stand out as the most preferable choices for OPTICS clustering, whereas the other dimensionality reduction methods offer smaller but still positive average gains over no-reduction baseline.

\subsection{Aggregate Analysis}\label{subsec:agg_analysis}
In this section, we summarize the impact of dimensionality reduction methods on the performance of clustering algorithms using two complementary statistics: (i) Percentage of wins that quantifies how often a reduced representation beats no-reduction baseline, and (ii) Average win/loss percentage that quantifies the average relative gain (or loss) in ARI compared to baseline. Tables \ref{tab:kmeans_aggregate} to \ref{tab:optics_aggregate} report these statistics for both synthetic and real-world datasets at the three dimensionality reduction levels considered. This aggregate view highlights consistent trends of improvement or degradation that are often obscured in boxplots. Figures \ref{fig:synthetic_max_gain} and \ref{fig:real_max_gain} in \ref{sec:apx_a} further report the largest average ARI gains over no reduction baseline along with their statistical significance, obtained for all pairings.

For $k$-means (see Table \ref{tab:kmeans_aggregate}), Kernel PCA wins over baseline on a majority of synthetic datasets, with strong average gains, indicating that nonlinear geometry can substantially improve centroid separability. However, this advantage largely fades on real-world data, with win rates of 50–60\% and only small positive ARI averages. In contrast, Isomap remains steadier across data types, with small positive averages on synthetic data. Isomap is competitive on real-world data as well, providing the highest win rate (65\% at the $k-1$ reduction level) among all competing methods. MDS shows the weakest performance, being strongly negative on synthetic data at the $k-1$ dimensionality reduction level and only around baseline on real-world data. VAE remains close to no reduction baseline on both synthetic and real-world data, with small average win/loss percentages at all levels, while PCA loses on average on synthetic data and stays at the baseline level on real-world data.

For hierarchical clustering (see Table \ref{tab:agglomerative_aggregate}), Kernel PCA clearly emerges as the most suitable dimensionality reduction method winning frequently on both synthetic (with 75.3\% of wins when original data are reduced to $k-1$ dimensions) and real-world (with 80.0\% of wins still when original data are reduced to $k-1$ dimensions) data, along with a considerable ARI improvement on synthetic data (36.3\% at the $k-1$ reduction level) and a smaller but clearly positive ARI improvement on real-world data (11.4\% at the $k-1$ reduction level). Isomap ranks second on synthetic data, delivering strong gains but more modest positive average changes on real data, whereas VAE ranks second on real-world data with average gains of about 5\% and win rates of 60–70\%. PCA provides a considerable average gain at the $k-1$ reduction level on synthetic data (19.2\%) and moderate positive changes on real data. By contrast, MDS is inconsistent — showing mixed results on both synthetic and real data. Finally, VAE provides positive averages at all levels on both synthetic and real-world data, but it wins on less than half of the synthetic datasets.

For GMM (see Table \ref{tab:gmm_aggregate}), synthetic data results show some benefits from nonlinear reductions. Kernel PCA yields the largest average gains (12–13\%) but wins on less than half of the datasets (47–48\%), while Isomap wins slightly more often (up to 52.9\% at the 50\% dimensionality reduction level) with smaller average gains (up to 6.5\%). However, on real-world data, these advantages disappear. Kernel PCA records negative averages at all levels. PCA with $k-1$ and 50\% reductions provides more wins than losses but yields negative win/loss ARI averages. Isomap, VAE, and MDS underperform on average as well, with negative averages at all levels. In practice, dimensionality reduction prior to GMM on real-world data offers little systematic benefit, underscoring the need for its cautious or dataset-specific application.

For OPTICS (see Table \ref{tab:optics_aggregate}), Kernel PCA, Isomap, and VAE provide substantial average ARI gains over the no-reduction baseline at all the three dimensionality reduction levels on synthetic data, while PCA and MDS only improve mean ARI at the $k-1$ reduction level. However, these average gains do not generally translate into improvements on a majority of datasets: only Kernel PCA at the $k-1$ reduction level achieves a win rate above 50\%. On real-world data, Kernel PCA provides the clearest gains at the $k-1$ and 25\% dimensionality reduction levels (with win percentages of 75.0 and 70.0\%, respectively). VAE also achieves high win rates on real data (with 70.0\% win percentage at the $k-1$ and 25\% reduction levels), along with positive win/loss ARI averages at all reduction levels. MDS, PCA, and Isomap also achieve positive averages on real-world data, but often show win percentages under the 50\% mark. Overall, OPTICS differs from GMM in that nonlinear reductions — particularly Kernel PCA and VAE — can yield systematic benefits on real data.

The results of all our experiments, including the source code and data, are available in our GitHub repository.

\begin{table}[H]
\centering
\caption{Comparison of performance improvements over no-reduction baseline for $k$-means clustering across different dimensionality reduction methods on synthetic and real-world datasets. Percentage of wins and average win/loss percentage in terms of ARI are reported.}
\label{tab:kmeans_aggregate}
{
\resizebox{\textwidth}{!}{
\begin{tabular}{l l cc cc}
\toprule
\textbf{Method} & \textbf{Reduction} & \multicolumn{2}{c}{\textbf{Percentage of wins}} & \multicolumn{2}{c}{\textbf{Average win (+) / loss (-) percentage}} \\
\cmidrule(lr){3-4} \cmidrule(lr){5-6}
& & \textbf{Synthetic data} & \textbf{Real-world data} & \textbf{Synthetic data} & \textbf{Real-world data} \\
\midrule
\multirow{3}{*}{PCA} & $k-1$ & 26.2 & 25.0 & -2.29 & -0.02 \\
 & 25\% & 19.7 & 20.0 & -2.63 & +0.56 \\
 & 50\% & 18.2 & 35.0 & -1.07 & -0.06 \\
\cmidrule(lr){1-6}
\multirow{3}{*}{Kernel PCA} & $k-1$ & 69.8 & 60.0 & +15.14 & +1.95 \\
 & 25\% & 67.9 & 50.0 & +14.45 & +0.84 \\
 & 50\% & 73.5 & 60.0 & +16.68 & +1.34 \\
\cmidrule(lr){1-6}
\multirow{3}{*}{VAE} & $k-1$ & 47.6 & 50.0 & -0.66 & +1.18 \\
 & 25\% & 49.1 & 55.0 & +0.92 & +1.44 \\
 & 50\% & 48.9 & 50.0 & +1.14 & +1.17 \\
\cmidrule(lr){1-6}
\multirow{3}{*}{Isomap} & $k-1$ & 52.2 & 65.0 & +1.04 & +2.08 \\
 & 25\% & 47.8 & 55.0 & +0.02 & +0.49 \\
 & 50\% & 50.4 & 60.0 & +0.82 & +1.90 \\
\cmidrule(lr){1-6}
\multirow{3}{*}{MDS} & $k-1$ & 28.8 & 35.0 & -10.07 & -0.32 \\
 & 25\% & 20.6 & 50.0 & -3.37 & -0.35 \\
 & 50\% & 20.4 & 35.0 & -1.14 & +0.17 \\
\bottomrule
\end{tabular}
}}
\end{table}

\begin{table}[H]
\centering
\caption{Comparison of performance improvements over no-reduction baseline for AHC clustering across different dimensionality reduction methods on synthetic and real-world datasets. Percentage of wins and average win/loss percentage in terms of ARI are reported.}
\label{tab:agglomerative_aggregate}
{\resizebox{\textwidth}{!}{
\begin{tabular}{l l cc cc}
\toprule
\textbf{Method} & \textbf{Reduction} & \multicolumn{2}{c}{\textbf{Percentage of wins}} & \multicolumn{2}{c}{\textbf{Average win (+) / loss (-) percentage}} \\
\cmidrule(lr){3-4} \cmidrule(lr){5-6}
& & \textbf{Synthetic data} & \textbf{Real-world data} & \textbf{Synthetic data} & \textbf{Real-world data} \\
\midrule
\multirow{3}{*}{PCA} & $k-1$ & 53.8 & 50.0 & +19.18 & +3.12 \\
 & 25\% & 32.8 & 30.0 & +0.33 & +2.27 \\
 & 50\% & 31.6 & 20.0 & +0.11 & +0.88 \\
\cmidrule(lr){1-6}
\multirow{3}{*}{Kernel PCA} & $k-1$ & 75.3 & 80.0 & +36.32 & +11.40 \\
 & 25\% & 66.8 & 70.0 & +30.35 & +10.70 \\
 & 50\% & 68.1 & 70.0 & +30.79 & +11.34 \\
\cmidrule(lr){1-6}
\multirow{3}{*}{VAE} & $k-1$ & 48.1 & 60.0 & +13.89 & +4.91 \\
 & 25\% & 44.3 & 60.0 & +9.84 & +4.66 \\
 & 50\% & 41.2 & 70.0 & +7.82 & +5.06 \\
\cmidrule(lr){1-6}
\multirow{3}{*}{Isomap} & $k-1$ & 55.3 & 60.0 & +21.17 & +2.72 \\
 & 25\% & 48.7 & 60.0 & +18.48 & +4.05 \\
 & 50\% & 49.5 & 50.0 & +18.12 & +2.31 \\
\cmidrule(lr){1-6}
\multirow{3}{*}{MDS} & $k-1$ & 53.0 & 35.0 & +10.17 & +1.76 \\
 & 25\% & 31.9 & 30.0 & -0.09 & +0.50 \\
 & 50\% & 32.2 & 30.0 & +0.19 & +0.75 \\
\bottomrule
\end{tabular}
}}
\end{table}

\begin{table}[H]
\centering
\caption{Comparison of performance improvements over no-reduction baseline for GMM clustering across different dimensionality reduction methods on synthetic and real-world datasets. Percentage of wins and average win/loss percentage in terms of ARI are reported.}
\label{tab:gmm_aggregate}
{\resizebox{\textwidth}{!}{
\begin{tabular}{l l cc cc}
\toprule
\textbf{Method} & \textbf{Reduction} & \multicolumn{2}{c}{\textbf{Percentage of wins}} & \multicolumn{2}{c}{\textbf{Average win (+) / loss (-) percentage}} \\
\cmidrule(lr){3-4} \cmidrule(lr){5-6}
& & \textbf{Synthetic data} & \textbf{Real-world data} & \textbf{Synthetic data} & \textbf{Real-world data} \\
\midrule
\multirow{3}{*}{PCA} & $k-1$ & 36.1 & 60.0 & -0.33 & -4.56 \\
 & 25\% & 31.1 & 45.0 & -2.26 & -4.95 \\
 & 50\% & 31.2 & 65.0 & -0.70 & -4.87 \\
\cmidrule(lr){1-6}
\multirow{3}{*}{Kernel PCA} & $k-1$ & 47.0 & 40.0 & +12.26 & -4.31 \\
 & 25\% & 48.4 & 40.0 & +12.47 & -5.61 \\
 & 50\% & 47.1 & 45.0 & +13.28 & -3.38 \\
\cmidrule(lr){1-6}
\multirow{3}{*}{VAE} & $k-1$ & 41.1 & 45.0 & -4.55 & -5.06 \\
 & 25\% & 36.5 & 45.0 & -2.54 & -2.91 \\
 & 50\% & 41.5 & 50.0 & -0.98 & -2.73 \\
\cmidrule(lr){1-6}
\multirow{3}{*}{Isomap} & $k-1$ & 50.6 & 35.0 & -0.24 & -4.70 \\
 & 25\% & 48.6 & 45.0 & +3.37 & -4.01 \\
 & 50\% & 52.9 & 50.0 & +6.54 & -4.14 \\
\cmidrule(lr){1-6}
\multirow{3}{*}{MDS} & $k-1$ & 24.1 & 40.0 & -9.43 & -4.46 \\
 & 25\% & 28.5 & 55.0 & -2.39 & -4.68 \\
 & 50\% & 31.7 & 45.0 & -1.57 & -4.56 \\
\bottomrule
\end{tabular}
}}
\end{table}

\begin{table}[H]
\centering
\caption{Comparison of performance improvements over no-reduction baseline for OPTICS clustering across different dimensionality reduction methods on synthetic and real-world datasets. Percentage of wins and average win/loss percentage in terms of ARI are reported.}
\label{tab:optics_aggregate}
{
\resizebox{\textwidth}{!}{
\begin{tabular}{l l cc cc}
\toprule
\textbf{Method} & \textbf{Reduction} & \multicolumn{2}{c}{\textbf{Percentage of wins}} & \multicolumn{2}{c}{\textbf{Average win (+) / loss (-) percentage}} \\
\cmidrule(lr){3-4} \cmidrule(lr){5-6}
& & \textbf{Synthetic data} & \textbf{Real-world data} & \textbf{Synthetic data} & \textbf{Real-world data} \\
\midrule
\multirow{3}{*}{PCA} & $k-1$ & 31.0 & 55.0 & +9.41 & +10.66 \\
 & 25\% & 7.6 & 40.0 & -0.48 & +9.58 \\
 & 50\% & 5.8 & 35.0 & -0.81 & +9.69 \\
\cmidrule(lr){1-6}
\multirow{3}{*}{Kernel PCA} & $k-1$ & 53.8 & 75.0 & +26.59 & +15.27 \\
 & 25\% & 44.4 & 70.0 & +19.51 & +10.71 \\
 & 50\% & 34.8 & 55.0 & +15.26 & +10.42 \\
\cmidrule(lr){1-6}
\multirow{3}{*}{VAE} & $k-1$ & 43.4 & 70.0 & +20.48 & +14.28 \\
 & 25\% & 40.4 & 70.0 & +18.97 & +9.34 \\
 & 50\% & 40.0 & 45.0 & +18.11 & +8.65 \\
\cmidrule(lr){1-6}
\multirow{3}{*}{Isomap} & $k-1$ & 49.7 & 60.0 & +23.44 & +12.10 \\
 & 25\% & 44.4 & 45.0 & +15.52 & +6.44 \\
 & 50\% & 39.0 & 55.0 & +12.15 & +9.54 \\
\cmidrule(lr){1-6}
\multirow{3}{*}{MDS} & $k-1$ & 34.2 & 35.0 & +14.17 & +9.45 \\
 & 25\% & 7.0 & 45.0 & -1.08 & +10.61 \\
 & 50\% & 6.1 & 25.0 & -0.45 & +2.82 \\
\bottomrule
\end{tabular}
}}
\end{table}

\section{Discussion}\label{sec:discussion}
Clustering results obtained on synthetic data demonstrate a marked influence of the selected dimensionality reduction method on each clustering algorithm considered. Nonlinear manifold learning methods, particularly Isomap and Kernel PCA, which preserve global and local geometry better than purely linear projections when the intrinsic data structure is nonlinear, consistently enhance the performance of the $k$-means, AHC, and OPTICS clustering algorithms, but provide smaller and less consistent gains when applied with GMM clustering. Importantly, the most suitable dimensionality reduction level clearly depends on the selected dimensionality reduction method/clustering algorithm pairing: Extreme reduction (to $k-1$ features) provides the largest gains for AHC and OPTICS with all methods, while a moderate reduction (to 50\% of the original number of features) is preferable for $k$-means and GMM with most methods, suggesting a balance required between noise removal and information preservation in order to obtain a good clustering. The VAE-based approach, while theoretically appealing for complex manifolds, shows instability, especially in the case of mixture model clustering, where it significantly degrades the results at the $k-1$ and 25\% dimensionality reduction levels, which is likely due to its probabilistic latent representation that introduces distortions in inter-point distances. MDS displays a high performance variation, excelling on some datasets with well-separated clusters but failing on others with overlapping manifolds, which reflects its sensitivity to noise and scaling. Importantly, consistent failure of some dimensionality reduction method/clustering algorithm pairings, such as PCA and MDS with $k$-means, VAE with GMM at the $k-1$ and 25\% dimensionality reduction levels, and MDS with GMM at $k-1$ dimensions, compared to no-reduction baseline (see Table \ref{tab:Wilcoxon_synthetic}) challenges the assumption that dimensionality reduction should always be performed prior to clustering. However, the success of Isomap and Kernel PCA supports the integration of dimensionality reduction — after careful data structure and clustering approach examination — as a standard preprocessing step in complex high-dimensional clustering pipelines.

The results obtained on real-world data suggest using a careful selection of the dimensionality reduction method/clustering algorithm pairing as well as of the dimensionality reduction level, with the aggressive reduction to $k-1$ dimensions often providing the largest gains (e.g., for AHC with Kernel PCA). 
We need to highlight that such a careful selection is necessary because dimensionality reduction can substantially lower average ARI scores relative to no-reduction baseline, whereas statistically significant improvements are limited to certain combinations of clustering algorithms, dimensionality reduction methods, and dimensionality reduction levels (see Tables \ref{tab:gmm_aggregate} and \ref{tab:Wilcoxon}). For hierarchical clustering (i.e., AHC), geometry-aware maps such as Kernel PCA with a well-tuned kernel and Isomap, as well as VAE, most reliably improve cluster separability (e.g., see the results for the Wine dataset in Table \ref{tab:agglomerative_real}), whereas for probabilistic clustering (i.e., GMM) no method improves over no-reduction baseline on average, despite some dataset-specific gains (e.g., the Wine dataset, see Table \ref{tab:gaussian_real}). $K$-means benefits from dimensionality reduction more opportunistically: Linear PCA or manifold learners can yield some ARI improvements on intrinsically low-dimensional structures (e.g., for the Iris and Wine datasets), but can also be neutral when clusters are already roughly spherical in the input space (see Table \ref{tab:kmeans_real}). OPTICS is uniquely fragile in the original high-dimensional space, where it often fails to detect any plausible cluster structure (see Table \ref{tab:optics_real}). Consequently, it benefits the most from dimensionality reduction. Kernel PCA and Isomap often succeed (e.g., on the Ecoli and Wine datasets), while VAE’s stochastic latent geometry can be useful as well (e.g., see the results obtained on the Ecoli, Segmentation, and Vertebral column datasets). When the obtained ARI scores are averaged over all clustering algorithms and reduction levels, Kernel PCA, VAE, and Isomap significantly outperform no-reduction baseline on real-world data (see Table \ref{tab:reducer_comparison}), while GMM and $k$-means are the best-performing clustering algorithms overall (see Table \ref{tab:clusterer_comparison}).

By aggregating the results obtained for synthetic and real-world data, the following practical recommendations can be provided: (i) 
Selecting the dimensionality reduction level according to the reduction method/clustering algorithm pairing, rather than favoring moderate or aggressive reduction in all cases, (ii) Preferring Kernel PCA and Isomap with hierarchical clustering, while considering no reduction baseline as a strong alternative option for mixture models, 
(iii) Using Kernel PCA and Isomap as strong, low-risk benchmarks with $k$-means, and (iv) Preferring Kernel PCA, Isomap, and VAE with OPTICS, while validating the VAE embeddings with neighborhood and density diagnostics prior to applying density-based clustering.
Our findings corroborate the conclusion that the domain matters. Thus, the main trends observed on synthetic data were not always noticeable on much more heterogeneous real-world benchmarks; for instance, the systematic gains observed for GMM on synthetic manifolds were not found on their real-world counterparts.


\section{Conclusion}\label{sec:conclusion}
We conducted a systematic study to provide statistically grounded answers to three consequential questions in unsupervised machine learning: 
\begin{itemize}
    \item Should we perform dimensionality reduction of raw data before clustering?
    \item Which dimensionality reduction methods can be recommended and which should be avoided for the clustering algorithm and dataset under consideration? ?
    \item How much can clustering performance be improved by using dimensionality reduction?
\end{itemize} 

We demonstrated that dimensionality reduction should be applied judiciously as its benefit hinges on the fit between the obtained embedding geometry and the downstream clustering algorithm. Three main conclusions emerge: (i) Dimensionality reduction method/clustering algorithm pairing is critical. These general trends are characteristic for both synthetic and real-world data – Kernel PCA is the most reliable enhancer for hierarchical and density-based clustering, 
nonlinear Isomap manifolds are particularly effective for k-means and hierarchical clustering, and for most types of clustering the PCA performance at the 25–50\% dimensionality reduction levels is very similar to that of no-reduction baseline (the sole exception is OPTICS' performance on real-world data). When bad pairings are used, dimensionality reduction can lead to a substantial decline in clustering performance.
(ii) 
The most suitable dimensionality reduction level clearly depends on the selected dimensionality reduction method/clustering algorithm pairing. An aggressive compression to $k-1$ dimensions provides the largest gains for hierarchical and density-based clustering, while keeping 50\% of features is generally safer for $k$-means and GMM;
(iii) Domain matters. Improvement attenuates on heterogeneous real-world datasets compared to their synthetic counterparts, especially in the case of $k$-means and GMM clustering. 


Some open avenues we would like to highlight are the following: Automated model selection for kernels and neighborhood graphs, joint learning of embeddings and clusters, evaluation of the methods robustness to noise and class imbalance, 
and the investigation of the applicability of deep clustering techniques, such as Deep Embedded Clustering (DEC) \cite{xie2016unsupervised} and Variational Deep Embedding (VaDE) \cite{jiang2016variational}. All of them deserve an additional systematic study under a comparable protocol. Our work aims to catalyze such efforts, helping practitioners select 
dimensionality reduction strategies that are tailored to their data geometry and clustering objectives.

\section*{Data availability}
All processed data used in this study along with the code needed to replicate our experiments are freely available in our GitHub repository at: \url{https://github.com/ELounissi/Dimensionality_Reduction_Clustering/}.


\newpage
\begin{landscape}

\appendix
\section{Detailed ARI Scores for Synthetic and Real-World Datasets} \label{sec:apx_a}

This appendix presents complete ARI results for all clustering algorithms ($k$-means, AHC, GMM, and OPTICS) applied to synthetic and real-world datasets, with and without dimensionality reduction. The initial number of dimensions, $D$, was reduced to $k-1$ dimensions (where $k$ is the known number of clusters), and 25\% and 50\% of $D$.

\renewcommand{\thetable}{A.\arabic{table}}
\setcounter{table}{0}
\begin{table}[H]
\centering
\caption{Mean ARI across 300 Circles datasets, with equal weight per dataset. The best ARI score in each row is highlighted in bold.}
\label{tab:ari_circles}
{
\resizebox{1.5\textwidth}{!}{
\begin{tabular}{l c ccc ccc ccc ccc ccc}
\toprule
& \textbf{No Reduction}
& \multicolumn{3}{c}{\textbf{PCA}} 
& \multicolumn{3}{c}{\textbf{Kernel PCA}} 
& \multicolumn{3}{c}{\textbf{VAE}} 
& \multicolumn{3}{c}{\textbf{Isomap}} 
& \multicolumn{3}{c}{\textbf{MDS}} \\

\cmidrule(lr){3-5}
\cmidrule(lr){6-8}
\cmidrule(lr){9-11}
\cmidrule(lr){12-14}
\cmidrule(lr){15-17}

\textbf{Algorithms} & 
& \textbf{$k-1$} & \textbf{25\%} & \textbf{50\%}
& \textbf{$k-1$} & \textbf{25\%} & \textbf{50\%}
& \textbf{$k-1$} & \textbf{25\%} & \textbf{50\%}
& \textbf{$k-1$} & \textbf{25\%} & \textbf{50\%}
& \textbf{$k-1$} & \textbf{25\%} & \textbf{50\%} \\

\midrule
$k$-means & 0.158 & 0.158 & 0.157 & 0.158 & 0.581 & 0.590 & \textbf{0.622} & 0.161 & 0.181 & 0.160 & 0.280 & 0.301 & 0.304 & 0.157 & 0.156 & 0.157 \\
AHC & 0.215 & 0.212 & 0.211 & 0.218 & \textbf{0.639} & 0.633 & 0.637 & 0.212 & 0.225 & 0.218 & 0.270 & 0.290 & 0.272 & 0.212 & 0.209 & 0.216 \\
GMM & 0.049 & 0.155 & 0.153 & 0.121 & 0.572 & 0.594 & \textbf{0.626} & 0.138 & 0.179 & 0.181 & 0.323 & 0.512 & 0.516 & 0.154 & 0.152 & 0.120 \\
OPTICS & 0.631 & 0.697 & 0.609 & 0.605 & 0.767 & 0.758 & \textbf{0.771} & 0.723 & 0.727 & 0.720 & 0.724 & 0.741 & 0.744 & 0.654 & 0.597 & 0.611 \\
\bottomrule
\end{tabular}
}

}
\end{table}

\begin{table}[H]
\centering
\caption{Mean ARI across 300 Moons datasets, with equal weight per dataset. The best ARI score in each row is highlighted in bold.}
\label{tab:ari_moons}
{
\resizebox{1.5\textwidth}{!}{
\begin{tabular}{l c ccc ccc ccc ccc ccc}
\toprule
& \textbf{No Reduction}
& \multicolumn{3}{c}{\textbf{PCA}}
& \multicolumn{3}{c}{\textbf{Kernel PCA}}
& \multicolumn{3}{c}{\textbf{VAE}}
& \multicolumn{3}{c}{\textbf{Isomap}}
& \multicolumn{3}{c}{\textbf{MDS}} \\

\cmidrule(lr){3-5}
\cmidrule(lr){6-8}
\cmidrule(lr){9-11}
\cmidrule(lr){12-14}
\cmidrule(lr){15-17}

\textbf{Algorithms} & 
& \textbf{$k-1$} & \textbf{25\%} & \textbf{50\%}
& \textbf{$k-1$} & \textbf{25\%} & \textbf{50\%}
& \textbf{$k-1$} & \textbf{25\%} & \textbf{50\%}
& \textbf{$k-1$} & \textbf{25\%} & \textbf{50\%}
& \textbf{$k-1$} & \textbf{25\%} & \textbf{50\%} \\

\midrule
$k$-means & 0.541 & 0.541 & 0.539 & 0.540 & 0.603 & 0.604 & 0.610 & 0.572 & 0.577 & 0.573 & \textbf{0.742} & 0.725 & 0.725 & 0.537 & 0.543 & 0.542 \\
AHC & 0.585 & 0.584 & 0.582 & 0.585 & 0.617 & 0.600 & 0.590 & 0.595 & 0.590 & 0.593 & \textbf{0.808} & 0.803 & 0.796 & 0.569 & 0.581 & 0.586 \\
GMM & 0.627 & 0.584 & 0.511 & 0.491 & 0.566 & 0.644 & 0.600 & 0.585 & 0.579 & 0.565 & \textbf{0.876} & 0.772 & 0.791 & 0.573 & 0.511 & 0.492 \\
OPTICS & 0.163 & 0.522 & 0.210 & 0.173 & 0.433 & 0.504 & 0.444 & 0.710 & 0.706 & 0.702 & \textbf{0.718} & 0.520 & 0.460 & 0.500 & 0.177 & 0.177 \\
\bottomrule
\end{tabular}
}

}
\end{table}
\end{landscape}

\begin{landscape}
\begin{table}[H]
\centering
\caption{Mean ARI across 265 Rodriguez Structured Gaussian datasets, with equal weight per dataset. The best ARI score in each row is highlighted in bold.}
\label{tab:ari_rodriguez}
{
\resizebox{1.5\textwidth}{!}{
\begin{tabular}{l c ccc ccc ccc ccc ccc}
\toprule
& \textbf{No Reduction}
& \multicolumn{3}{c}{\textbf{PCA}}
& \multicolumn{3}{c}{\textbf{Kernel PCA}}
& \multicolumn{3}{c}{\textbf{VAE}}
& \multicolumn{3}{c}{\textbf{Isomap}}
& \multicolumn{3}{c}{\textbf{MDS}} \\

\cmidrule(lr){3-5}
\cmidrule(lr){6-8}
\cmidrule(lr){9-11}
\cmidrule(lr){12-14}
\cmidrule(lr){15-17}

\textbf{Algorithms} & & \textbf{$k-1$} & \textbf{25\%} & \textbf{50\%}
& \textbf{$k-1$} & \textbf{25\%} & \textbf{50\%}
& \textbf{$k-1$} & \textbf{25\%} & \textbf{50\%}
& \textbf{$k-1$} & \textbf{25\%} & \textbf{50\%}
& \textbf{$k-1$} & \textbf{25\%} & \textbf{50\%} \\

\midrule
$k$-means & 0.625 & 0.529 & 0.527 & 0.579 & 0.658 & 0.627 & 0.671 & 0.553 & 0.583 & 0.617 & \textbf{0.848} & 0.804 & 0.832 & 0.629 & 0.558 & 0.593 \\
AHC & 0.523 & 0.548 & 0.391 & 0.459 & 0.674 & 0.552 & 0.608 & 0.496 & 0.436 & 0.465 & \textbf{0.840} & 0.747 & 0.777 & 0.641 & 0.438 & 0.477 \\
GMM & 0.810 & 0.729 & 0.730 & \textbf{0.847} & 0.810 & 0.722 & 0.762 & 0.569 & 0.616 & 0.695 & 0.778 & 0.790 & 0.825 & 0.731 & 0.712 & 0.807 \\
OPTICS & 0.298 & 0.117 & 0.247 & 0.277 & 0.231 & 0.198 & 0.225 & 0.224 & 0.269 & 0.272 & \textbf{0.435} & 0.359 & 0.318 & 0.200 & 0.264 & 0.285 \\
\bottomrule
\end{tabular}
}

}
\end{table}

\begin{table}[H]
\centering
\caption{Mean ARI across 300 Repliclust datasets, with equal weight per dataset. The best ARI score in each row is highlighted in bold.}
\label{tab:ari_repliclust}
{
\resizebox{1.5\textwidth}{!}{
\begin{tabular}{l c ccc ccc ccc ccc ccc}
\toprule
 & \textbf{No Reduction}
 & \multicolumn{3}{c}{\textbf{PCA}}
 & \multicolumn{3}{c}{\textbf{Kernel PCA}}
 & \multicolumn{3}{c}{\textbf{VAE}}
 & \multicolumn{3}{c}{\textbf{Isomap}}
 & \multicolumn{3}{c}{\textbf{MDS}} \\
 
\cmidrule(lr){3-5}
\cmidrule(lr){6-8}
\cmidrule(lr){9-11}
\cmidrule(lr){12-14}
\cmidrule(lr){15-17}

\textbf{Algorithms} & & \textbf{$k-1$} & \textbf{25\%} & \textbf{50\%}
& \textbf{$k-1$} & \textbf{25\%} & \textbf{50\%}
& \textbf{$k-1$} & \textbf{25\%} & \textbf{50\%}
& \textbf{$k-1$} & \textbf{25\%} & \textbf{50\%}
& \textbf{$k-1$} & \textbf{25\%} & \textbf{50\%} \\
\midrule
$k$-means & 0.912 & 0.909 & 0.899 & 0.912 & 0.986 & 0.978 & \textbf{0.986} & 0.916 & 0.926 & 0.930 & 0.433 & 0.428 & 0.431 & 0.522 & 0.840 & 0.897 \\
AHC & 0.162 & 0.889 & 0.299 & 0.221 & \textbf{0.983} & 0.882 & 0.856 & 0.717 & 0.607 & 0.506 & 0.426 & 0.389 & 0.374 & 0.472 & 0.244 & 0.208 \\
GMM & 0.974 & 0.968 & 0.969 & 0.978 & \textbf{0.987} & 0.974 & 0.981 & 0.962 & 0.964 & 0.966 & 0.469 & 0.514 & 0.583 & 0.626 & 0.980 & 0.978 \\
OPTICS & 0.002 & 0.103 & 0.004 & 0.004 & \textbf{0.688} & 0.380 & 0.238 & 0.224 & 0.126 & 0.101 & 0.143 & 0.083 & 0.046 & 0.280 & 0.010 & 0.003 \\
\bottomrule
\end{tabular}
}

}
\end{table}
\end{landscape}

\begin{landscape}
\begin{table}[H]
\centering
\caption{
ARI scores for $k$-means on 20 real-world datasets from the UCI repository, with results reported for each dimensionality reduction method and reduction level. The best ARI score for each dataset is highlighted in bold.}
\label{tab:kmeans_real}
{
\resizebox{1.5\textwidth}{!}{
\begin{tabular}{l c ccc ccc ccc ccc ccc}
\toprule
& \textbf{No Reduction}
& \multicolumn{3}{c}{\textbf{PCA}}
& \multicolumn{3}{c}{\textbf{Kernel PCA}}
& \multicolumn{3}{c}{\textbf{VAE}}
& \multicolumn{3}{c}{\textbf{Isomap}}
& \multicolumn{3}{c}{\textbf{MDS}} \\

\cmidrule(lr){3-5}
\cmidrule(lr){6-8}
\cmidrule(lr){9-11}
\cmidrule(lr){12-14}
\cmidrule(lr){15-17}

\textbf{Dataset} & & \textbf{$k-1$} & \textbf{25\%} & \textbf{50\%}
& \textbf{$k-1$} & \textbf{25\%} & \textbf{50\%}
& \textbf{$k-1$} & \textbf{25\%} & \textbf{50\%}
& \textbf{$k-1$} & \textbf{25\%} & \textbf{50\%}
& \textbf{$k-1$} & \textbf{25\%} & \textbf{50\%} \\

\midrule
Breast tissue & 0.269 & 0.272 & 0.264 & 0.272 & 0.275 & \textbf{0.339} & 0.275 & 0.266 & 0.270 & 0.266 & 0.298 & 0.299 & 0.298 & 0.260 & 0.288 & 0.260 \\
Breast Wisconsin & 0.671 & 0.659 & 0.671 & 0.671 & 0.655 & 0.643 & 0.649 & 0.713 & 0.728 & 0.696 & 0.753 & \textbf{0.755} & \textbf{0.755} & 0.689 & 0.677 & 0.671 \\
Ecoli & 0.502 & 0.506 & 0.459 & 0.482 & 0.416 & 0.434 & 0.363 & 0.481 & 0.446 & 0.487 & \textbf{0.716} & 0.443 & 0.708 & 0.503 & 0.366 & 0.521 \\
Glass & 0.169 & 0.170 & \textbf{0.266} & 0.170 & 0.177 & 0.178 & 0.177 & 0.188 & 0.227 & 0.188 & 0.141 & 0.140 & 0.141 & 0.165 & 0.171 & 0.165 \\
Haberman & -0.001 & 0.013 & -0.003 & 0.013 & -0.007 & 0.001 & -0.007 & \textbf{0.053} & 0.035 & \textbf{0.053} & 0.014 & -0.014 & 0.014 & 0.008 & -0.002 & 0.008 \\
Ionosphere & 0.168 & 0.158 & 0.168 & 0.168 & 0.208 & \textbf{0.213} & \textbf{0.213} & 0.146 & 0.172 & 0.163 & 0.136 & 0.136 & 0.136 & 0.113 & 0.168 & 0.168 \\
Iris & 0.620 & 0.620 & \textbf{0.802} & 0.620 & 0.641 & 0.559 & 0.641 & 0.678 & 0.694 & 0.678 & 0.612 & 0.641 & 0.612 & 0.610 & 0.702 & 0.610 \\
Movement libras & 0.316 & 0.310 & 0.317 & 0.315 & \textbf{0.341} & 0.332 & 0.326 & 0.297 & 0.293 & 0.304 & 0.297 & 0.293 & 0.297 & 0.318 & 0.316 & 0.315 \\
Musk & 0.004 & 0.001 & 0.004 & 0.004 & 0.001 & -0.000 & -0.000 & -0.001 & 0.005 & 0.005 & \textbf{0.007} & \textbf{0.007} & \textbf{0.007} & -0.003 & 0.004 & 0.004 \\
Parkinsons & -0.098 & -0.098 & -0.098 & -0.098 & \textbf{0.161} & \textbf{0.161} & \textbf{0.161} & 0.063 & 0.077 & 0.069 & 0.094 & 0.094 & 0.094 & -0.074 & -0.091 & -0.098 \\
Segmentation & 0.461 & 0.452 & 0.453 & 0.463 & 0.508 & 0.461 & \textbf{0.514} & 0.461 & 0.450 & 0.431 & 0.392 & 0.394 & 0.382 & 0.461 & 0.465 & 0.463 \\
Sonar all & -0.002 & -0.002 & -0.002 & -0.002 & 0.019 & 0.016 & 0.016 & \textbf{0.025} & 0.006 & 0.006 & -0.005 & -0.004 & -0.004 & -0.004 & -0.002 & -0.001 \\
Spectf & -0.103 & -0.103 & -0.103 & -0.103 & \textbf{-0.054} & -0.079 & -0.082 & -0.097 & -0.103 & -0.105 & -0.094 & -0.097 & -0.096 & -0.105 & -0.103 & -0.103 \\
Transfusion & 0.053 & 0.051 & 0.049 & 0.051 & 0.013 & 0.003 & 0.013 & 0.031 & 0.037 & 0.031 & 0.062 & \textbf{0.068} & 0.062 & 0.032 & 0.020 & 0.032 \\
Vehicle & 0.076 & 0.076 & 0.076 & 0.075 & 0.075 & 0.074 & 0.072 & 0.077 & 0.076 & 0.075 & 0.089 & \textbf{0.096} & 0.094 & 0.080 & 0.077 & 0.076 \\
Vertebral column & 0.211 & 0.241 & 0.241 & 0.248 & 0.255 & 0.255 & 0.238 & 0.245 & 0.245 & 0.246 & 0.248 & 0.248 & \textbf{0.256} & 0.254 & 0.254 & 0.249 \\
Vowel context & 0.167 & 0.165 & 0.158 & 0.177 & 0.184 & 0.150 & 0.177 & 0.164 & 0.160 & 0.168 & 0.177 & 0.171 & \textbf{0.193} & 0.165 & 0.159 & 0.174 \\
Wine & 0.897 & 0.895 & 0.847 & 0.897 & 0.949 & 0.915 & \textbf{0.949} & 0.868 & 0.891 & 0.895 & 0.847 & 0.848 & 0.848 & 0.852 & 0.864 & 0.896 \\
Wine quality red & 0.062 & 0.057 & 0.059 & 0.058 & 0.047 & 0.059 & 0.059 & 0.068 & 0.068 & 0.069 & 0.065 & \textbf{0.081} & 0.067 & 0.059 & 0.070 & 0.060 \\
Yeast & 0.166 & 0.163 & 0.094 & 0.115 & 0.136 & 0.065 & 0.124 & 0.121 & 0.121 & 0.120 & \textbf{0.178} & 0.109 & 0.126 & 0.163 & 0.139 & 0.176 \\
\bottomrule
\end{tabular}
}

}
\end{table}
\end{landscape}

\begin{landscape}
\begin{table}[H]
\centering
\caption{ARI scores for AHC on 20 real-world datasets from the UCI repository, with results reported for each dimensionality reduction method and reduction level. The best ARI score for each dataset is highlighted in bold.}
\label{tab:agglomerative_real}
{
\resizebox{1.5\textwidth}{!}{
\begin{tabular}{l c ccc ccc ccc ccc ccc}
\toprule
& \textbf{No Reduction}
& \multicolumn{3}{c}{\textbf{PCA}}
& \multicolumn{3}{c}{\textbf{Kernel PCA}}
& \multicolumn{3}{c}{\textbf{VAE}}
& \multicolumn{3}{c}{\textbf{Isomap}}
& \multicolumn{3}{c}{\textbf{MDS}} \\

\cmidrule(lr){3-5}
\cmidrule(lr){6-8}
\cmidrule(lr){9-11}
\cmidrule(lr){12-14}
\cmidrule(lr){15-17}

\textbf{Dataset} & & \textbf{$k-1$} & \textbf{25\%} & \textbf{50\%}
& \textbf{$k-1$} & \textbf{25\%} & \textbf{50\%}
& \textbf{$k-1$} & \textbf{25\%} & \textbf{50\%}
& \textbf{$k-1$} & \textbf{25\%} & \textbf{50\%}
& \textbf{$k-1$} & \textbf{25\%} & \textbf{50\%} \\

\midrule
Breast tissue & 0.156 & 0.156 & 0.280 & 0.156 & 0.267 & \textbf{0.345} & 0.267 & 0.223 & 0.213 & 0.223 & 0.158 & 0.241 & 0.158 & 0.265 & 0.274 & 0.265 \\
Breast Wisconsin & 0.005 & 0.002 & 0.005 & 0.005 & \textbf{0.713} & 0.672 & 0.660 & 0.002 & 0.000 & 0.000 & 0.005 & 0.005 & 0.005 & 0.005 & 0.005 & 0.005 \\
Ecoli & 0.726 & \textbf{0.746} & 0.691 & 0.665 & 0.430 & 0.501 & 0.517 & 0.689 & 0.607 & 0.701 & 0.664 & 0.686 & 0.713 & 0.722 & 0.607 & 0.719 \\
Glass & 0.021 & \textbf{0.256} & 0.017 & \textbf{0.256} & 0.200 & 0.206 & 0.200 & 0.186 & 0.178 & 0.186 & 0.066 & 0.066 & 0.066 & 0.023 & 0.009 & 0.023 \\
Haberman & \textbf{0.012} & \textbf{0.012} & 0.007 & \textbf{0.012} & -0.006 & -0.004 & -0.006 & 0.010 & 0.007 & 0.010 & \textbf{0.012} & \textbf{0.012} & \textbf{0.012} & \textbf{0.012} & 0.007 & \textbf{0.012} \\
Ionosphere & 0.004 & 0.096 & 0.004 & 0.004 & 0.192 & \textbf{0.288} & 0.264 & 0.057 & 0.099 & 0.075 & 0.132 & 0.004 & 0.004 & 0.173 & 0.004 & 0.004 \\
Iris & 0.558 & 0.558 & \textbf{0.758} & 0.558 & 0.641 & 0.549 & 0.641 & 0.682 & 0.688 & 0.682 & 0.564 & 0.702 & 0.564 & 0.558 & 0.702 & 0.558 \\
Movement libras & 0.314 & 0.299 & 0.314 & 0.314 & \textbf{0.372} & 0.322 & 0.329 & 0.309 & 0.292 & 0.305 & 0.299 & 0.278 & 0.306 & 0.312 & 0.314 & 0.314 \\
Musk & -0.002 & 0.003 & -0.002 & -0.002 & 0.002 & 0.004 & -0.002 & -0.002 & -0.003 & -0.002 & \textbf{0.008} & 0.003 & -0.002 & -0.002 & -0.002 & -0.002 \\
Parkinsons & -0.026 & -0.026 & -0.026 & -0.026 & \textbf{0.161} & 0.118 & 0.142 & -0.020 & -0.021 & -0.019 & -0.007 & -0.013 & -0.026 & -0.026 & -0.026 & -0.026 \\
Segmentation & 0.000 & 0.000 & 0.000 & 0.000 & \textbf{0.484} & 0.449 & 0.412 & 0.459 & 0.464 & 0.458 & 0.213 & 0.219 & 0.208 & 0.000 & 0.000 & 0.000 \\
Sonar all & -0.001 & -0.001 & -0.001 & -0.001 & \textbf{0.013} & -0.004 & -0.004 & -0.001 & -0.001 & -0.001 & -0.005 & -0.001 & -0.001 & -0.001 & -0.001 & -0.001 \\
Spectf & \textbf{-0.006} & -0.054 & \textbf{-0.006} & \textbf{-0.006} & -0.099 & -0.090 & -0.092 & -0.027 & -0.031 & -0.036 & -0.058 & \textbf{-0.006} & \textbf{-0.006} & \textbf{-0.006} & \textbf{-0.006} & \textbf{-0.006} \\
Transfusion & \textbf{0.031} & 0.020 & \textbf{0.031} & 0.020 & 0.025 & -0.000 & 0.025 & -0.004 & 0.010 & -0.004 & \textbf{0.031} & \textbf{0.031} & \textbf{0.031} & \textbf{0.031} & -0.004 & \textbf{0.031} \\
Vehicle & 0.001 & 0.001 & 0.001 & 0.001 & 0.099 & 0.108 & 0.106 & 0.031 & 0.049 & 0.044 & 0.091 & 0.101 & \textbf{0.110} & 0.001 & 0.001 & 0.001 \\
Vertebral column & -0.004 & 0.230 & 0.230 & -0.007 & 0.228 & 0.228 & \textbf{0.250} & 0.090 & 0.090 & 0.094 & 0.021 & 0.021 & 0.030 & -0.004 & -0.004 & -0.004 \\
Vowel context & 0.147 & 0.154 & 0.114 & 0.161 & 0.181 & 0.147 & \textbf{0.191} & 0.166 & 0.160 & 0.159 & 0.139 & 0.152 & 0.158 & 0.144 & 0.148 & 0.148 \\
Wine & 0.790 & 0.896 & 0.770 & 0.787 & 0.931 & 0.896 & \textbf{0.931} & 0.853 & 0.855 & 0.858 & 0.880 & 0.915 & 0.800 & 0.869 & 0.798 & 0.832 \\
Wine quality red & -0.002 & 0.002 & 0.002 & 0.002 & 0.057 & 0.057 & 0.049 & 0.002 & 0.001 & 0.002 & 0.052 & \textbf{0.117} & 0.055 & 0.002 & -0.001 & 0.002 \\
Yeast & 0.010 & 0.010 & -0.002 & 0.010 & \textbf{0.124} & 0.085 & 0.123 & 0.011 & 0.010 & 0.011 & 0.012 & 0.012 & 0.012 & 0.010 & 0.009 & 0.010 \\
\bottomrule
\end{tabular}
}

}
\end{table}
\end{landscape}



\begin{landscape}
\begin{table}[H]
\centering
\caption{ARI scores for GMM on 20 real-world datasets from the UCI repository, with results reported for each dimensionality reduction method and reduction level. The best ARI score for each dataset is highlighted in bold.}
\label{tab:gaussian_real}
{
\resizebox{1.5\textwidth}{!}{
\begin{tabular}{l c ccc ccc ccc ccc ccc}
\toprule
& \textbf{No Reduction}
& \multicolumn{3}{c}{\textbf{PCA}}
& \multicolumn{3}{c}{\textbf{Kernel PCA}}
& \multicolumn{3}{c}{\textbf{VAE}}
& \multicolumn{3}{c}{\textbf{Isomap}}
& \multicolumn{3}{c}{\textbf{MDS}} \\

\cmidrule(lr){3-5}
\cmidrule(lr){6-8}
\cmidrule(lr){9-11}
\cmidrule(lr){12-14}
\cmidrule(lr){15-17}

\textbf{Dataset } & & \textbf{$k-1$} & \textbf{25\%} & \textbf{50\%}
& \textbf{$k-1$} & \textbf{25\%} & \textbf{50\%}
& \textbf{$k-1$} & \textbf{25\%} & \textbf{50\%}
& \textbf{$k-1$} & \textbf{25\%} & \textbf{50\%}
& \textbf{$k-1$} & \textbf{25\%} & \textbf{50\%} \\

\midrule
Breast tissue & 0.276 & 0.297 & 0.273 & 0.297 & 0.308 & \textbf{0.345} & 0.308 & 0.296 & 0.319 & 0.296 & 0.284 & 0.322 & 0.284 & 0.306 & 0.325 & 0.306 \\
Breast Wisconsin & \textbf{0.774} & 0.610 & 0.465 & 0.077 & 0.649 & 0.666 & 0.707 & 0.537 & 0.586 & 0.641 & 0.684 & 0.452 & 0.520 & 0.582 & 0.377 & 0.077 \\
Ecoli & 0.633 & 0.647 & 0.476 & 0.662 & 0.396 & 0.455 & 0.404 & 0.499 & 0.533 & 0.541 & 0.477 & 0.656 & 0.585 & 0.579 & 0.651 & \textbf{0.732} \\
Glass & 0.153 & 0.191 & \textbf{0.204} & 0.191 & 0.164 & 0.153 & 0.164 & 0.194 & 0.194 & 0.194 & 0.172 & 0.161 & 0.172 & 0.190 & 0.186 & 0.190 \\
Haberman & 0.104 & \textbf{0.135} & -0.003 & \textbf{0.135} & -0.007 & 0.002 & -0.007 & 0.099 & 0.063 & 0.099 & 0.032 & -0.023 & 0.032 & 0.074 & -0.002 & 0.074 \\
Ionosphere & \textbf{0.394} & 0.022 & 0.307 & 0.300 & 0.197 & 0.005 & 0.193 & -0.003 & 0.221 & 0.192 & 0.082 & 0.188 & 0.336 & 0.051 & 0.313 & 0.276 \\
Iris & \textbf{0.904} & 0.577 & 0.757 & 0.577 & 0.630 & 0.505 & 0.630 & 0.640 & 0.689 & 0.640 & 0.533 & 0.647 & 0.533 & 0.644 & 0.669 & 0.644 \\
Movement libras & 0.349 & 0.381 & 0.360 & 0.329 & 0.344 & 0.333 & 0.319 & 0.317 & 0.308 & 0.317 & 0.298 & 0.305 & 0.312 & \textbf{0.412} & 0.357 & 0.328 \\
Musk & 0.005 & 0.006 & -0.004 & \textbf{0.087} & 0.004 & -0.001 & -0.000 & 0.004 & 0.002 & 0.004 & 0.006 & 0.008 & 0.007 & -0.001 & -0.004 & \textbf{0.087} \\
Parkinsons & -0.086 & -0.097 & -0.053 & -0.037 & 0.169 & 0.129 & 0.156 & 0.051 & 0.067 & 0.082 & 0.229 & \textbf{0.230} & 0.065 & -0.051 & -0.061 & -0.027 \\
Segmentation & 0.412 & 0.418 & 0.415 & 0.388 & 0.458 & 0.457 & 0.460 & 0.477 & 0.459 & \textbf{0.487} & 0.363 & 0.327 & 0.390 & 0.419 & 0.432 & 0.391 \\
Sonar all & -0.001 & 0.003 & 0.001 & -0.000 & 0.013 & \textbf{0.021} & -0.003 & 0.009 & 0.015 & 0.008 & 0.019 & -0.004 & 0.006 & -0.002 & 0.002 & -0.003 \\
Spectf & -0.103 & -0.100 & -0.101 & -0.093 & -0.105 & -0.076 & -0.081 & -0.101 & -0.087 & -0.093 & -0.106 & -0.079 & \textbf{-0.063} & -0.103 & -0.094 & -0.086 \\
Transfusion & -0.048 & 0.054 & 0.058 & 0.054 & 0.054 & -0.001 & 0.054 & 0.020 & 0.041 & 0.020 & \textbf{0.060} & 0.055 & \textbf{0.060} & 0.032 & 0.010 & 0.032 \\
Vehicle & 0.093 & 0.062 & 0.067 & 0.086 & 0.080 & 0.059 & 0.072 & 0.065 & 0.088 & 0.082 & 0.076 & 0.090 & \textbf{0.104} & 0.074 & 0.072 & 0.087 \\
Vertebral column & \textbf{0.539} & 0.257 & 0.257 & 0.272 & 0.237 & 0.237 & 0.222 & 0.255 & 0.255 & 0.275 & 0.247 & 0.247 & 0.268 & 0.299 & 0.299 & 0.295 \\
Vowel context & 0.179 & 0.173 & 0.153 & 0.180 & 0.180 & 0.155 & \textbf{0.195} & 0.178 & 0.170 & 0.186 & 0.170 & 0.135 & 0.186 & 0.168 & 0.156 & 0.172 \\
Wine & 0.910 & 0.898 & 0.769 & 0.931 & 0.765 & 0.898 & \textbf{0.933} & 0.860 & 0.880 & 0.871 & 0.817 & 0.899 & 0.774 & 0.867 & 0.752 & 0.862 \\
Wine quality red & 0.057 & 0.066 & 0.063 & 0.063 & 0.051 & 0.060 & 0.049 & 0.061 & 0.066 & 0.076 & 0.053 & 0.062 & 0.055 & \textbf{0.089} & 0.069 & 0.060 \\
Yeast & 0.042 & 0.072 & 0.132 & 0.114 & 0.135 & 0.063 & 0.134 & 0.117 & 0.133 & 0.121 & 0.151 & 0.105 & 0.128 & 0.066 & 0.144 & \textbf{0.177} \\
\bottomrule
\end{tabular}
}

}
\end{table}
\end{landscape}

\begin{landscape}
\begin{table}[H]
\centering
\caption{ARI scores for OPTICS on 20 real-world datasets from the UCI repository, with results reported for each dimensionality reduction method and reduction level. The best ARI score for each dataset is highlighted in bold.}
\label{tab:optics_real}
{
\resizebox{1.5\textwidth}{!}{
\begin{tabular}{l c ccc ccc ccc ccc ccc}
\toprule
& \textbf{No Reduction}
& \multicolumn{3}{c}{\textbf{PCA}}
& \multicolumn{3}{c}{\textbf{Kernel PCA}}
& \multicolumn{3}{c}{\textbf{VAE}}
& \multicolumn{3}{c}{\textbf{Isomap}}
& \multicolumn{3}{c}{\textbf{MDS}} \\

\cmidrule(lr){3-5}
\cmidrule(lr){6-8}
\cmidrule(lr){9-11}
\cmidrule(lr){12-14}
\cmidrule(lr){15-17}

\textbf{Dataset} & & \textbf{$k-1$} & \textbf{25\%} & \textbf{50\%}
& \textbf{$k-1$} & \textbf{25\%} & \textbf{50\%}
& \textbf{$k-1$} & \textbf{25\%} & \textbf{50\%}
& \textbf{$k-1$} & \textbf{25\%} & \textbf{50\%}
& \textbf{$k-1$} & \textbf{25\%} & \textbf{50\%} \\

\midrule
Breast tissue & 0.332 & 0.329 & 0.234 & 0.329 & 0.286 & 0.366 & 0.286 & 0.321 & 0.333 & 0.321 & \textbf{0.380} & 0.378 & \textbf{0.380} & 0.330 & 0.313 & 0.330 \\
Breast Wisconsin & 0.000 & 0.694 & 0.000 & 0.000 & 0.638 & 0.000 & 0.000 & 0.716 & 0.000 & 0.000 & 0.523 & 0.000 & 0.000 & \textbf{0.761} & 0.000 & 0.000 \\
Ecoli & 0.000 & 0.392 & 0.666 & 0.681 & \textbf{0.689} & 0.646 & 0.675 & 0.637 & 0.579 & 0.636 & 0.648 & 0.463 & 0.498 & 0.000 & 0.685 & 0.000 \\
Glass & 0.170 & 0.156 & 0.182 & 0.156 & \textbf{0.218} & 0.190 & \textbf{0.218} & 0.141 & 0.177 & 0.141 & 0.104 & 0.118 & 0.104 & 0.128 & 0.198 & 0.128 \\
Haberman & -0.010 & 0.011 & 0.006 & 0.011 & -0.010 & -0.003 & -0.010 & -0.008 & 0.008 & -0.008 & \textbf{0.015} & -0.018 & \textbf{0.015} & 0.000 & 0.001 & 0.000 \\
Ionosphere & 0.041 & 0.141 & 0.041 & 0.064 & 0.213 & 0.202 & \textbf{0.241} & 0.131 & 0.021 & 0.023 & 0.150 & 0.000 & 0.000 & 0.129 & 0.070 & 0.064 \\
Iris & 0.450 & 0.669 & \textbf{0.787} & 0.669 & 0.640 & 0.543 & 0.640 & 0.667 & 0.670 & 0.667 & 0.568 & 0.568 & 0.568 & 0.633 & 0.642 & 0.633 \\
Movement libras & 0.051 & 0.100 & 0.030 & 0.051 & \textbf{0.284} & 0.171 & 0.110 & 0.180 & 0.176 & 0.159 & 0.131 & 0.131 & 0.126 & 0.030 & 0.030 & 0.051 \\
Musk & 0.000 & 0.002 & 0.000 & 0.000 & 0.000 & -0.000 & -0.004 & 0.001 & 0.003 & -0.000 & 0.005 & \textbf{0.010} & 0.000 & -0.001 & 0.000 & 0.000 \\
Parkinsons & \textbf{0.442} & 0.168 & 0.423 & \textbf{0.442} & 0.169 & 0.056 & 0.000 & 0.148 & 0.161 & 0.128 & 0.083 & 0.000 & 0.350 & 0.174 & 0.389 & \textbf{0.442} \\
Segmentation & 0.404 & 0.210 & 0.254 & 0.407 & 0.466 & 0.435 & \textbf{0.544} & 0.479 & 0.445 & 0.360 & 0.387 & 0.387 & 0.457 & 0.456 & 0.460 & 0.460 \\
Sonar all & 0.000 & 0.019 & 0.122 & 0.000 & 0.011 & 0.011 & -0.001 & 0.055 & 0.035 & 0.002 & -0.005 & 0.000 & 0.000 & -0.002 & \textbf{0.129} & 0.000 \\
Spectf & 0.000 & \textbf{0.161} & 0.000 & 0.000 & -0.041 & 0.000 & 0.000 & 0.105 & 0.000 & 0.000 & -0.009 & 0.000 & 0.000 & 0.000 & 0.000 & 0.000 \\
Transfusion & 0.059 & -0.033 & -0.030 & -0.033 & 0.070 & 0.003 & 0.070 & 0.028 & -0.005 & 0.028 & \textbf{0.082} & -0.053 & \textbf{0.082} & 0.057 & 0.006 & 0.057 \\
Vehicle & 0.000 & 0.000 & 0.000 & 0.000 & 0.109 & \textbf{0.114} & 0.000 & 0.047 & 0.051 & 0.029 & 0.107 & 0.005 & 0.111 & 0.000 & 0.000 & 0.000 \\
Vertebral column & 0.000 & 0.239 & 0.239 & 0.317 & 0.258 & 0.258 & 0.253 & 0.248 & 0.248 & 0.274 & 0.310 & 0.310 & 0.266 & \textbf{0.351} & \textbf{0.351} & 0.000 \\
Vowel context & 0.000 & 0.000 & 0.000 & 0.000 & 0.078 & \textbf{0.114} & 0.052 & 0.021 & 0.005 & 0.009 & 0.000 & 0.066 & 0.040 & 0.000 & 0.000 & 0.000 \\
Wine & 0.000 & 0.815 & 0.880 & 0.784 & 0.914 & 0.871 & \textbf{0.949} & 0.878 & 0.888 & 0.898 & 0.879 & 0.864 & 0.848 & 0.784 & 0.787 & 0.337 \\
Wine quality red & 0.000 & 0.000 & 0.000 & 0.000 & 0.000 & 0.000 & 0.000 & 0.000 & \textbf{0.012} & 0.000 & 0.000 & 0.000 & 0.000 & 0.000 & 0.000 & 0.000 \\
Yeast & 0.000 & 0.000 & 0.019 & 0.000 & 0.000 & \textbf{0.105} & 0.000 & 0.000 & 0.000 & 0.000 & 0.000 & 0.000 & 0.000 & 0.000 & 0.000 & 0.000 \\
\bottomrule
\end{tabular}
}

}
\end{table}
\end{landscape}

\begin{landscape}

\begin{table}[H]
\centering
\small
\caption{Wilcoxon signed-rank test results for synthetic data. One-sided test was performed, $H_1: ARI_{method} > ARI_{baseline}, \alpha = 0.05, n = 1165$. The hypothesis $H_1$ is that a given dimensionality reduction method significantly improves the ARI value (obtained after clustering) of no-reduction baseline; \textit{p}-values shown correspond to the ARI improvements compared to no-reduction baseline. Significant \textit{p}-values ($p \le 0.05$) are highlighted in bold.
}
\label{tab:Wilcoxon_synthetic}
{
\setlength{\tabcolsep}{0.5pt}
\renewcommand{\wilcoxonpvalue}[2]{\makebox[3em][c]{#1\makebox[0pt][l]{#2}}}
\begin{adjustbox}{max width=1.12\linewidth,center}
\begin{tabular}{lccccccccccccccc}

\toprule & \multicolumn{3}{c}{PCA} & \multicolumn{3}{c}{Kernel PCA} & \multicolumn{3}{c}{VAE} & \multicolumn{3}{c}{Isomap} & \multicolumn{3}{c}{MDS} \\ \cmidrule(lr){2-4} \cmidrule(lr){5-7} \cmidrule(lr){8-10} \cmidrule(lr){11-13} \cmidrule(lr){14-16} Algorithm/Reduction & $k-1$ & 25\% & 50\% & $k-1$ & 25\% & 50\% & $k-1$ & 25\% & 50\% & $k-1$ & 25\% & 50\% & $k-1$ & 25\% & 50\% \\ \midrule $k$-means & \wilcoxonpvalue{0.999}{} & \wilcoxonpvalue{0.999}{} & \wilcoxonpvalue{0.992}{} & \wilcoxonpvalue{\textbf{0.021}}{} & \wilcoxonpvalue{0.082}{} & \wilcoxonpvalue{\textbf{0.015}}{} & \wilcoxonpvalue{0.930}{} & \wilcoxonpvalue{0.716}{} & \wilcoxonpvalue{0.522}{} & \wilcoxonpvalue{\textbf{0.003}}{} & \wilcoxonpvalue{\textbf{0.014}}{} & \wilcoxonpvalue{\textbf{0.005}}{} & \wilcoxonpvalue{0.972}{} & \wilcoxonpvalue{0.996}{} & \wilcoxonpvalue{0.996}{} \\ AHC & \wilcoxonpvalue{\textbf{0.035}}{} & \wilcoxonpvalue{0.984}{} & \wilcoxonpvalue{0.872}{} & \wilcoxonpvalue{\textbf{0.001}}{} & \wilcoxonpvalue{\textbf{0.032}}{} & \wilcoxonpvalue{\textbf{0.005}}{} & \wilcoxonpvalue{0.277}{} & \wilcoxonpvalue{0.523}{} & \wilcoxonpvalue{0.403}{} & \wilcoxonpvalue{\textbf{0.001}}{} & \wilcoxonpvalue{\textbf{0.003}}{} & \wilcoxonpvalue{\textbf{0.001}}{} & \wilcoxonpvalue{\textbf{0.002}}{} & \wilcoxonpvalue{0.914}{} & \wilcoxonpvalue{0.607}{} \\ GMM & \wilcoxonpvalue{0.727}{} & \wilcoxonpvalue{0.821}{} & \wilcoxonpvalue{0.624}{} & \wilcoxonpvalue{0.115}{} & \wilcoxonpvalue{0.402}{} & \wilcoxonpvalue{0.505}{} & \wilcoxonpvalue{1.000}{} & \wilcoxonpvalue{0.987}{} & \wilcoxonpvalue{0.930}{} & \wilcoxonpvalue{0.351}{} & \wilcoxonpvalue{0.387}{} & \wilcoxonpvalue{0.223}{} & \wilcoxonpvalue{0.980}{} & \wilcoxonpvalue{0.891}{} & \wilcoxonpvalue{0.949}{} \\ OPTICS & \wilcoxonpvalue{0.482}{} & \wilcoxonpvalue{0.430}{} & \wilcoxonpvalue{0.757}{} & \wilcoxonpvalue{\textbf{0.016}}{} & \wilcoxonpvalue{\textbf{0.027}}{} & \wilcoxonpvalue{\textbf{0.035}}{} & \wilcoxonpvalue{0.058}{} & \wilcoxonpvalue{\textbf{0.004}}{} & \wilcoxonpvalue{\textbf{0.008}}{} & \wilcoxonpvalue{\textbf{0.001}}{} & \wilcoxonpvalue{\textbf{0.002}}{} & \wilcoxonpvalue{\textbf{0.033}}{} & \wilcoxonpvalue{0.236}{} & \wilcoxonpvalue{0.545}{} & \wilcoxonpvalue{0.259}{} \\ \bottomrule \end{tabular} \end{adjustbox} } \end{table}

\begin{table}[H]
\centering
\caption{Wilcoxon signed-rank test results for 20 real-world benchmarks. One-sided test was performed, $H_1: ARI_{method} > ARI_{baseline}, \alpha = 0.05, n = 20$. The hypothesis $H_1$ is that a given dimensionality reduction method significantly improves the ARI value (obtained after clustering) of no-reduction baseline; \textit{p}-values shown correspond to the ARI improvements compared to no-reduction baseline. Significant \textit{p}-values ($p \le 0.05$) are highlighted in bold.}
\label{tab:Wilcoxon}
{
\small
\setlength{\tabcolsep}{0.5pt}
\renewcommand{\wilcoxonpvalue}[2]{\makebox[3em][c]{#1\makebox[0pt][l]{#2}}}
\begin{adjustbox}{max width=1.12\linewidth,center}
\begin{tabular}{lccccccccccccccc}
\toprule
 & \multicolumn{3}{c}{PCA} & \multicolumn{3}{c}{Kernel PCA} & \multicolumn{3}{c}{VAE} & \multicolumn{3}{c}{Isomap} & \multicolumn{3}{c}{MDS} \\
\cmidrule(lr){2-4} \cmidrule(lr){5-7} \cmidrule(lr){8-10} \cmidrule(lr){11-13} \cmidrule(lr){14-16}
Algorithm/Reduction & $k-1$ & 25\% & 50\% & $k-1$ & 25\% & 50\% & $k-1$ & 25\% & 50\% & $k-1$ & 25\% & 50\% & $k-1$ & 25\% & 50\% \\
\midrule
$k$-means & \wilcoxonpvalue{0.917}{} & \wilcoxonpvalue{0.910}{} & \wilcoxonpvalue{0.424}{} & \wilcoxonpvalue{0.062}{} & \wilcoxonpvalue{0.420}{} & \wilcoxonpvalue{0.101}{} & \wilcoxonpvalue{0.249}{} & \wilcoxonpvalue{0.205}{} & \wilcoxonpvalue{0.311}{} & \wilcoxonpvalue{0.156}{} & \wilcoxonpvalue{0.522}{} & \wilcoxonpvalue{0.249}{} & \wilcoxonpvalue{0.826}{} & \wilcoxonpvalue{0.390}{} & \wilcoxonpvalue{0.515}{} \\
AHC & \wilcoxonpvalue{0.053}{} & \wilcoxonpvalue{0.500}{} & \wilcoxonpvalue{0.663}{} & \wilcoxonpvalue{\textbf{0.003}}{\textbf{\boldmath\,}} & \wilcoxonpvalue{\textbf{0.013}}{\textbf{\boldmath\,}} & \wilcoxonpvalue{\textbf{0.005}}{\textbf{\boldmath\,}} & \wilcoxonpvalue{\textbf{0.027}}{\textbf{\boldmath\,}} & \wilcoxonpvalue{0.057}{} & \wilcoxonpvalue{\textbf{0.022}}{\textbf{\boldmath\,}} & \wilcoxonpvalue{\textbf{0.036}}{\textbf{\boldmath\,}} & \wilcoxonpvalue{\textbf{0.004}}{\textbf{\boldmath\,}} & \wilcoxonpvalue{\textbf{0.016}}{\textbf{\boldmath\,}} & \wilcoxonpvalue{0.168}{} & \wilcoxonpvalue{0.414}{} & \wilcoxonpvalue{0.074}{} \\
GMM & \wilcoxonpvalue{0.522}{} & \wilcoxonpvalue{0.929}{} & \wilcoxonpvalue{0.324}{} & \wilcoxonpvalue{0.869}{} & \wilcoxonpvalue{0.853}{} & \wilcoxonpvalue{0.715}{} & \wilcoxonpvalue{0.774}{} & \wilcoxonpvalue{0.727}{} & \wilcoxonpvalue{0.663}{} & \wilcoxonpvalue{0.918}{} & \wilcoxonpvalue{0.806}{} & \wilcoxonpvalue{0.806}{} & \wilcoxonpvalue{0.785}{} & \wilcoxonpvalue{0.774}{} & \wilcoxonpvalue{0.622}{} \\
OPTICS & \wilcoxonpvalue{\textbf{0.046}}{\textbf{\boldmath\,}} & \wilcoxonpvalue{0.160}{} & \wilcoxonpvalue{0.072}{} & \wilcoxonpvalue{\textbf{0.004}}{\textbf{\boldmath\,}} & \wilcoxonpvalue{\textbf{0.004}}{\textbf{\boldmath\,}} & \wilcoxonpvalue{\textbf{0.020}}{\textbf{\boldmath\,}} & \wilcoxonpvalue{\textbf{0.006}}{\textbf{\boldmath\,}} & \wilcoxonpvalue{\textbf{0.012}}{\textbf{\boldmath\,}} & \wilcoxonpvalue{0.238}{} & \wilcoxonpvalue{\textbf{0.016}}{\textbf{\boldmath\,}} & \wilcoxonpvalue{0.159}{} & \wilcoxonpvalue{\textbf{0.016}}{\textbf{\boldmath\,}} & \wilcoxonpvalue{0.241}{} & \wilcoxonpvalue{\textbf{0.042}}{\textbf{\boldmath\,}} & \wilcoxonpvalue{0.175}{} \\
\bottomrule
\end{tabular}
\end{adjustbox}
}
\end{table}

\clearpage

\begin{table}[H]
\centering
\caption{Wilcoxon signed-rank test results comparing pairwise ARI performances of four dimensionality reduction methods and no-reduction baseline on 20 real-world benchmarks. Two-sided test was performed: $H_1: ARI_{method} \neq ARI_{baseline}, \alpha = 0.05$. ARI scores were averaged over three dimensionality reduction levels and four clustering algorithms. Significant \textit{p}-values ($p \le 0.05$) are highlighted in bold, and the sign in cell (i, j) indicates whether dimensionality reduction method i provides significantly higher (+) or lower (-) ARI scores than dimensionality reduction method j.
}

\label{tab:reducer_comparison}
{
\setlength{\tabcolsep}{3pt}
\begin{adjustbox}{max width=\linewidth}
\begin{tabular}{lcccccc}
\toprule
 & No reduction & PCA & Kernel PCA & VAE & Isomap & MDS \\
\midrule
No reduction & \wilcoxonpvalue{--}{} & \wilcoxonpvalue{0.177}{} & \wilcoxonpvalue{\textbf{0.001}}{\textbf{\boldmath\,($-$)}} & \wilcoxonpvalue{\textbf{0.001}}{\textbf{\boldmath\,($-$)}} & \wilcoxonpvalue{\textbf{0.033}}{\textbf{\boldmath\,($-$)}} & \wilcoxonpvalue{0.202}{} \\
PCA & \wilcoxonpvalue{0.177}{} & \wilcoxonpvalue{--}{} & \wilcoxonpvalue{0.058}{} & \wilcoxonpvalue{0.105}{} & \wilcoxonpvalue{0.475}{} & \wilcoxonpvalue{0.452}{} \\
Kernel PCA & \wilcoxonpvalue{\textbf{0.001}}{\textbf{\boldmath\,($+$)}} & \wilcoxonpvalue{0.058}{} & \wilcoxonpvalue{--}{} & \wilcoxonpvalue{0.105}{} & \wilcoxonpvalue{\textbf{0.033}}{\textbf{\boldmath\,($+$)}} & \wilcoxonpvalue{\textbf{0.011}}{\textbf{\boldmath\,($+$)}} \\
VAE & \wilcoxonpvalue{\textbf{0.001}}{\textbf{\boldmath\,($+$)}} & \wilcoxonpvalue{0.105}{} & \wilcoxonpvalue{0.105}{} & \wilcoxonpvalue{--}{} & \wilcoxonpvalue{0.674}{} & \wilcoxonpvalue{\textbf{0.009}}{\textbf{\boldmath\,($+$)}} \\
Isomap & \wilcoxonpvalue{\textbf{0.033}}{\textbf{\boldmath\,($+$)}} & \wilcoxonpvalue{0.475}{} & \wilcoxonpvalue{\textbf{0.033}}{\textbf{\boldmath\,($-$)}} & \wilcoxonpvalue{0.674}{} & \wilcoxonpvalue{--}{} & \wilcoxonpvalue{0.154}{} \\
MDS & \wilcoxonpvalue{0.202}{} & \wilcoxonpvalue{0.452}{} & \wilcoxonpvalue{\textbf{0.011}}{\textbf{\boldmath\,($-$)}} & \wilcoxonpvalue{\textbf{0.009}}{\textbf{\boldmath\,($-$)}} & \wilcoxonpvalue{0.154}{} & \wilcoxonpvalue{--}{} \\
\bottomrule
\end{tabular}
\end{adjustbox}
}
\end{table}
\begin{table}[H]
\centering
\caption{Wilcoxon signed-rank test results comparing pairwise ARI performances of four clustering algorithms on 20 real-world benchmarks. Two-sided test was performed: $H_1: ARI_{method} \neq ARI_{baseline}, \alpha = 0.05$. ARI scores were averaged over three dimensionality reduction levels and six dimensionality reduction methods (including no reduction baseline). Significant \textit{p}-values ($p \le 0.05$) are highlighted in bold, and the sign in cell (i, j) indicates whether algorithm i provides significantly higher (+) or lower (-) average ARI scores than algorithm j.
}
\label{tab:clusterer_comparison}
{
\setlength{\tabcolsep}{3pt}
\begin{adjustbox}{max width=\linewidth}
\begin{tabular}{lcccc}
\toprule
 & $k$-means & AHC & GMM & OPTICS \\
\midrule
$k$-means & \wilcoxonpvalue{--}{} & \wilcoxonpvalue{\textbf{0.006}}{\textbf{\boldmath\,($+$)}} & \wilcoxonpvalue{0.330}{} & \wilcoxonpvalue{0.058}{} \\
AHC & \wilcoxonpvalue{\textbf{0.006}}{\textbf{\boldmath\,($-$)}} & \wilcoxonpvalue{--}{} & \wilcoxonpvalue{\textbf{0.003}}{\textbf{\boldmath\,($-$)}} & \wilcoxonpvalue{0.546}{} \\
GMM & \wilcoxonpvalue{0.330}{} & \wilcoxonpvalue{\textbf{0.003}}{\textbf{\boldmath\,($+$)}} & \wilcoxonpvalue{--}{} & \wilcoxonpvalue{\textbf{0.024}}{\textbf{\boldmath\,($+$)}} \\
OPTICS & \wilcoxonpvalue{0.058}{} & \wilcoxonpvalue{0.546}{} & \wilcoxonpvalue{\textbf{0.024}}{\textbf{\boldmath\,($-$)}} & \wilcoxonpvalue{--}{} \\
\bottomrule
\end{tabular}
\end{adjustbox}
}
\end{table}

\clearpage
\setcounter{figure}{0}
\renewcommand{\thefigure}{A.\arabic{figure}}

\begin{figure}[H]
\centering
{
\setlength{\fboxsep}{2pt}
\fbox{\includegraphics[width=0.84\linewidth]{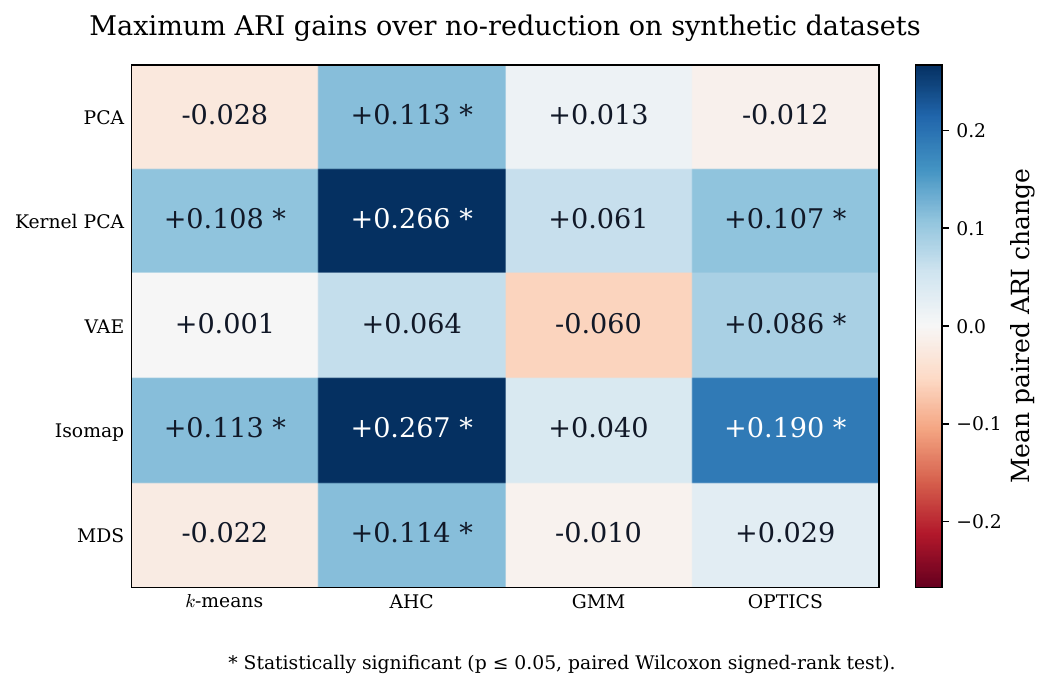}}
}
\caption{Best average ARI gains over no-reduction baseline for each dimensionality reduction method/clustering algorithm pairing obtained over three different dimensionality reduction levels for synthetic data. Each cell reports the average gain/loss provided by the pairing at its most favorable dimensionality reduction level, where the average is taken over all datasets. Positive values indicate the best improvements over no-reduction baseline, while negative values indicate the smallest losses (at the most favorable reduction level). Asterisks mark statistically significant differences with no-reduction baseline (according to the one-sided Wilcoxon signed-rank test, $p \le 0.05$; see Table~\ref{tab:Wilcoxon_synthetic}).}
\label{fig:synthetic_max_gain}
\end{figure}

\clearpage

\begin{figure}[H]
\centering
{
\setlength{\fboxsep}{2pt}
\fbox{\includegraphics[width=0.84\linewidth]{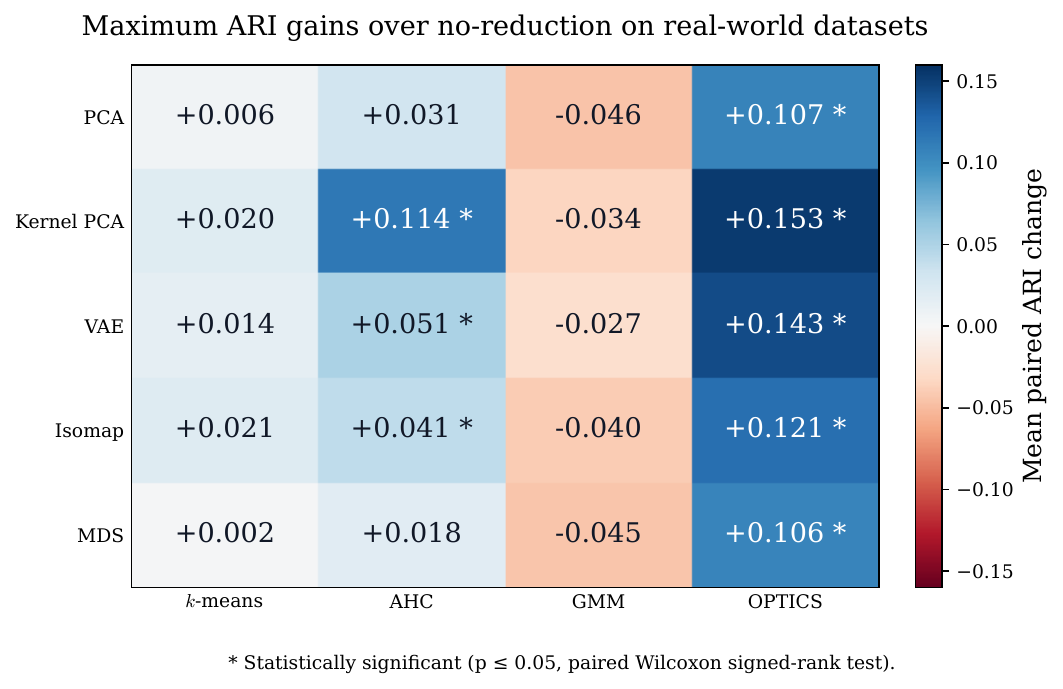}}
}
\caption{
    Best average ARI gains over no-reduction baseline for each dimensionality reduction method/clustering algorithm pairing obtained over three different dimensionality reduction levels for 20 real-world benchmarks. Each cell reports the average gain/loss provided by the pairing at its most favorable dimensionality reduction level, where the average is taken over all datasets. Positive values indicate the best improvements over no-reduction baseline, while negative values indicate the smallest losses (at the most favorable reduction level). Asterisks mark statistically significant differences with no-reduction baseline (according to the one-sided Wilcoxon signed-rank test, $p \le 0.05$; see Table~\ref{tab:Wilcoxon}).
    }
\label{fig:real_max_gain}
\end{figure}

\end{landscape}

\newpage
\bibliographystyle{elsarticle-num}
\bibliography{references}

\end{document}